\documentclass[sigconf]{acmart}

\AtBeginDocument{%
  }

\setcopyright{acmlicensed}
\copyrightyear{2025}
\acmYear{2025}
\acmDOI{3701551.3703577}

\acmConference[WSDM '25]{the 18th ACM International Conference on Web Search and Data Mining}{March 10-14, 2025}{Hannover, Germany}

\acmISBN{978-1-4503-XXXX-X/18/06}

\usepackage{xspace}
\usepackage{booktabs}
\usepackage{multirow}
\usepackage{enumitem}
\usepackage{titlesec}
\usepackage{enumerate}
\usepackage{placeins}
\usepackage{float}
\newcommand{\ours}{\textsc{TinyLLM}\xspace}

\begin{document}

\title{Beyond Answers: Transferring Reasoning Capabilities to Smaller LLMs Using Multi-Teacher Knowledge Distillation}

\author{Yijun Tian}
\authornote{Equally contributed.}
\email{yijun.tian@nd.edu}
\affiliation{%
  \institution{University of Notre Dame}
  \city{}
  \country{USA}
}

\author{Yikun Han}
\authornotemark[1]
\email{yikunhan@umich.edu}
\affiliation{%
  \institution{University of Michigan}
  \city{}
  \country{USA}}

\author{Xiusi Chen}
\authornotemark[1]
\email{xchen@cs.ucla.edu}
\affiliation{%
 \institution{University of California, Los Angeles}
 \city{}
 \country{USA}}

\author{Wei Wang}
\email{weiwang@cs.ucla.edu}
\affiliation{%
 \institution{University of California, Los Angeles}
 \city{}
 \country{USA}}

\author{Nitesh V. Chawla}
\email{nchawla@nd.edu}
\affiliation{%
  \institution{University of Notre Dame}
  \city{}
  \country{USA}
}

\renewcommand{\shortauthors}{Tian et al.}

\begin{abstract}
Transferring the reasoning capability from stronger large language models (LLMs) to smaller ones has been quite appealing, as smaller LLMs are more flexible to deploy with less expense. Among the existing solutions, knowledge distillation stands out due to its outstanding efficiency and generalization. However, existing methods suffer from several drawbacks, including limited knowledge diversity and the lack of rich contextual information. To solve the problems and facilitate the learning of compact language models, we propose \ours, a new knowledge distillation paradigm to learn a small student LLM from multiple large teacher LLMs. In particular, we encourage the student LLM to not only generate the correct answers but also understand the rationales behind these answers. Given that different LLMs possess diverse reasoning skills, we guide the student model to assimilate knowledge from various teacher LLMs. We further introduce an in-context example generator and a teacher-forcing Chain-of-Thought strategy to ensure that the rationales are accurate and grounded in contextually appropriate scenarios. Extensive experiments on six datasets across two reasoning tasks demonstrate the superiority of our method. Results show that \ours can outperform large teacher LLMs significantly, despite a considerably smaller model size. The source code is available at: \url{https://github.com/YikunHan42/TinyLLM}.
\end{abstract}



\maketitle

\section{Introduction}
Large language models (LLMs) have recently taken over various domains and web applications, including society~\cite{rao2023makes},
education~\cite{zelikman2023generating}, and recommendations \cite{wu2023survey}.
Although cutting-edge language models like GPT-4 and Claude-2 have shown remarkable capability in producing coherent and contextually appropriate text, their smaller counterparts often fall short, especially in tasks that demand sophisticated reasoning and a deep level of understanding \cite{wei2022emergentabilities}.
This discrepancy has been unveiled as the well-known scaling law of LLMs, which suggests a correlation between model size and reasoning, linguistic, and generalization capabilities \cite{kaplan2020scalinglaws}. 
However, deploying these colossal models in a real-world setting poses significant challenges due to their computational requirements and resource demands, underscoring the importance of building efficient, smaller models that retain the power of their larger counterparts.
Previous studies have shown that knowledge distillation is an instrumental tool in mitigating the performance gap between larger LLMs and smaller ones \cite{wan2023efficient, hsieh-etal-2023-distilling}. 
Examples of effective distillation methods include 
DistilBERT \cite{sanh2019distilbert}, Alpaca \cite{alpaca} and
Vicuna \cite{zheng2023judging}.

However, existing methods suffer from two major drawbacks: 
(1) \textbf{Limited Knowledge Diversity}: 
Current research predominantly employs a single-teacher approach, which confines the learning scope of the student model to the knowledge derived from its own training and architecture designs 
\cite{ho2022largelanguagemodels,magister2022teachingsmall,li2023symbolicchain,wang2022pinto}. 
This restricts the student model to a single perspective, potentially overlooking the diverse problem-solving strategies and reasoning capabilities exhibited by different models, limiting its breadth and depth of understanding.
(2) \textbf{Lack of Rich Contextual Information}: 
While rationales play a vital role in effective reasoning 
\cite{wei2022chainofthought,kojima2022zero}, current research primarily focuses on leveraging ground truth labels, which indicate the correct answer but do not provide insights into the reasoning and thought process behind that answer. In other words, learning the ground truth labels exclusively failed to capture the nuanced decision-making processes of the teachers, which are crucial for tasks requiring complex reasoning and interpretation. 

To solve these issues, we propose \ours, a paradigm that facilitates the learning of a small student LLM by distilling knowledge from multiple large teacher LLMs with rationale guidance. 
Specifically, \ours mitigates the limited knowledge diversity issue by involving multiple teacher models as \textit{co-advisors}, which introduces a richer, varied knowledge source for the student to learn from. To fully exploit each teacher model and mitigate the lack of rich contextual information problem, \ours asks the teacher for the credible rationales to support the answers, thereby providing the student with a deeper understanding of the problem-solving process. 
By learning from multiple teachers, the student model can inherit a broader range of skills and knowledge, leading to better generalization capabilities.
In addition, to ensure the rationales are grounded in contextually appropriate scenarios and reflect the true underlying reasoning procedure, \ours features an in-context example generator and a teacher-forcing Chain-of-Thought strategy, making the teachers understand the task through demonstrations and therefore generate the accurate rationales.

To thoroughly evaluate our approach, we conduct experiments on six datasets in commonsense and biomedical reasoning tasks. The results show that the usage of our paradigm enhances performance by \textbf{+5.07\%} to \textbf{+15.69\%} compared to full fine-tuning.
Compared to the teacher models, \ours achieves superior performance improvement, e.g., up to \textbf{+23.40\%} with significantly smaller model size, e.g., \textbf{1.1\%} to \textbf{26.0\%}.
In addition, compared to the state-of-the-art distillation methods, our approach improves the distillation performance by \textbf{+10.00\%} to \textbf{+11.79\%} across different model sizes. With the aim of further validating the effectiveness of our method, we perform efficiency analyses, ablation studies, parameter sensitivities, and case studies to provide a comprehensive evaluation across multiple dimensions. To summarize, our main contributions are as follows:

\begin{itemize}[nosep,leftmargin=*]
\item We identify two critical limitations in the existing knowledge distillation landscape for LLMs: 1) limited knowledge diversity, and 2) lack of rich contextual information.
\item To solve these two problems, we propose \ours, a novel knowledge distillation paradigm to learn a small student LLM from multiple large teacher LLMs. 
\item \ours encompasses several innovative designs including an in-context example generator, a teacher-forcing Chain-of-Thought strategy, and a joint learning objective from various teachers.
\item Extensive experiments validate the superiority of \ours across six datasets and two reasoning tasks, with performance improving by
up to \textbf{+15.69\%} compared to full fine-tuning, up to \textbf{+23.40\%} compared to teacher models, and up to \textbf{+11.79\%} compared to state-of-the-art.
In addition, \ours holds a significantly smaller model size, e.g., \textbf{1.1\%} to \textbf{26.0\%} compared to the teachers.
\end{itemize}

\section{Related Work}
In this section, we review existing work including large language models, chain of thought, and knowledge distillation.

\label{sec:Related Work}
\noindent
\textbf{Large Language Models.}
Recent advancements have seen the proposal of various Large Language Models (LLMs) \cite{raffel2020exploringlimits, touvron2023llama, touvron2023llama2, brown2020language, dubey2024llama}, which have showcased remarkable performance across a spectrum of tasks \cite{shi2023mededit, wei2024llmrec, prompt_llm_for_graph, oppu, wang2023hadskip, sku}. Central to these developments is question answering, a task that necessitates intricate reasoning and comprehensive understanding skills for text interpretation and generating suitable responses to queries \cite{lu2022learntoexplain, zhu2021retrievingreading, chen2023minprompt, gnp}. Despite their formidable learning capabilities, LLMs encounter limitations in accurately capturing factual knowledge and are prone to producing unsubstantiated responses \cite{zhao2023surveyllms, ji2023hallucinationnlg, bang2023multitaskmultilingual}. Moreover, the extensive number of parameters within LLMs complicates their adaptation for downstream tasks \cite{scao2022bloom, smith2022deepspeedmegatron}. To mitigate these challenges, several approaches aim to lessen the dependency on intensive training and reduce computational costs \cite{lester2021powerofscale, li2021prefixtuning, hu2022lora, wang2024roselora}. For example, Prompt Tuning \cite{lester2021powerofscale, liu2021p, gu2021ppt, wang2022no} employs soft prompts to adapt pre-trained LLMs for specific tasks. 

\noindent
\textbf{Chain of Thought.}
Recently, the use of rationales generated by LLMs has become a popular trend, setting itself apart from the traditional reliance on human-generated rationales \cite{hase2021learningfromexplanations, huang2022towards}.
Previously, human rationales have been used for model regularization \cite{ross2017rightreasons}, as additional inputs for predictions \cite{rajani2019explainyourself}, and to improve model performance \cite{zaidan2007annotatorrationales, zhang2016rationalecnn, camburu2018esnli, hancock2019feedchatbot, pruthi2022evaluatingexplanations, pmlr-v202-fu23d, huang2023large, zhangmultimodal}. They also serve as gold standard labels for generating similar rationales to enhance interpretability \cite{wiegreffe2021measuringassociation, narang2020wt5, eisenstein2022honeststudents, turpin2024language}. However, the cost of human rationales limits their widespread use.
On the other hand, modern LLMs can generate high-quality reasoning steps to explain their predictions \cite{wei2022chainofthought, kojima2022zero}, improving performance in few-shot or zero-shot learning \cite{wei2022chainofthought, kojima2022zero, wang2022pinto} and serving as self-improvement data \cite{zelikman2022star, huang2022languagemodelsselfimprove}. However, LLMs' size hinders their deployment in practice. Correspondingly, recent research explores leveraging generated rationales for training smaller, task-specific models with minimal computational and memory overhead \cite{wang2021reduce, ho2022largelanguagemodels, magister2022teachingsmall, li2023symbolicchain}. For example, PINTO \cite{wang2022pinto} presents an LLM pipeline that rationalizes via prompt-based learning.
However, they still rely on an LLM for rationale generation at test-time, not fully addressing deployment challenges.
In this work, we propose a multi-task learning paradigm with superior chain-of-thought reasoning capabilities, avoiding the dependence on teacher models during the test phase.

\begin{figure*}[htbp]
\begin{center}
\includegraphics[width=\linewidth]{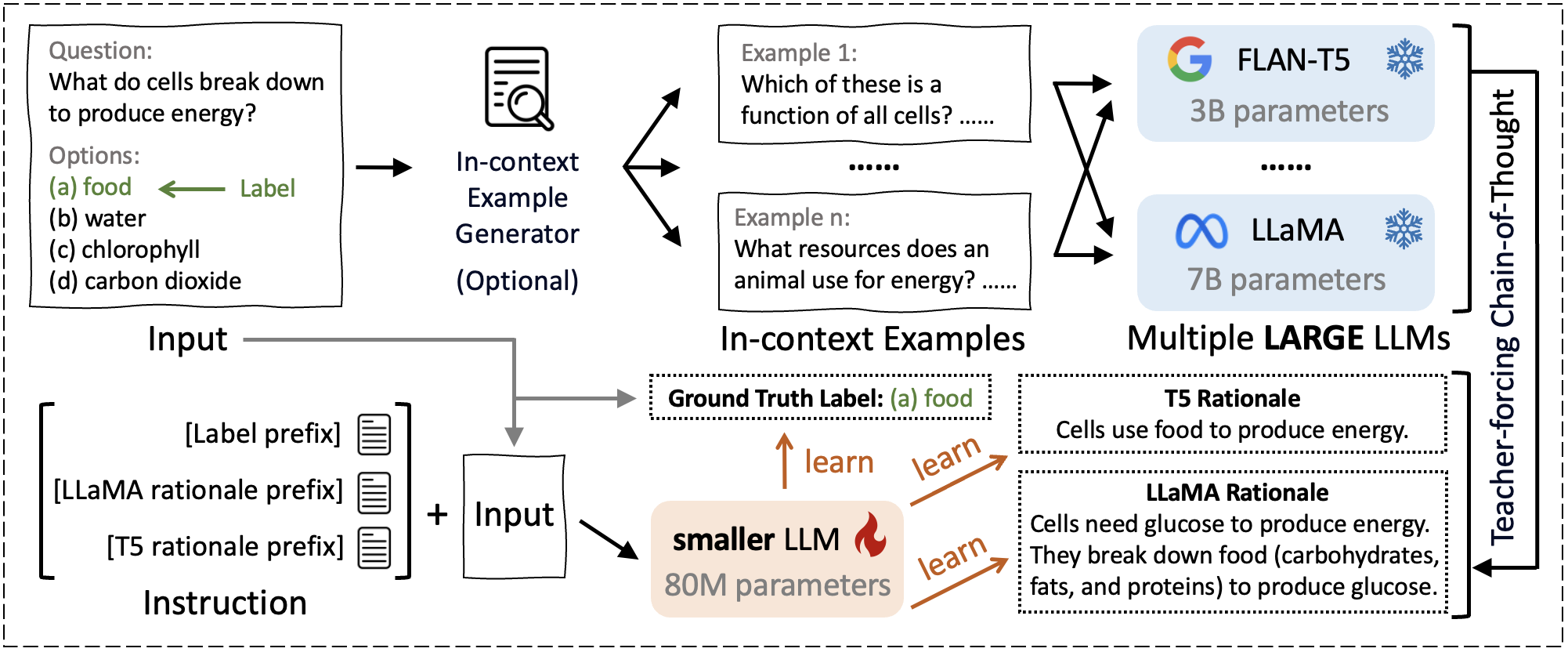}
\end{center}
\caption{
Pipeline of \ours: Given an input question, we first generate in-context examples and obtain rationales from multiple large LLMs via a teacher-forcing Chain-of-Thought strategy. 
Later, a small student LLM is trained to integrate rationales from different teachers via multi-task instruction tuning, along with the ground truth label.
}
\label{fig:pipeline}
\end{figure*}

\noindent
\textbf{Knowledge Distillation.}
Recent advancements in LLMs, such as PaLM 540B \cite{chowdhery2023palm, anil2023palm} and LLaMA 3.1 405B \cite{dubey2024llama}, bring formidable challenges in terms of inference and fine-tuning due to their immense computational requirements. Knowledge distillation \cite{hinton_kd, gou2021knowledge, nosmog, kd_on_graph_survey, cho2019efficacy, guo2023boosting, agarwal2024generalized, xu2024survey} has emerged as a pivotal approach to mitigate these challenges by training smaller student models to mimic the behavior of larger teacher models, significantly reducing resource demands. White-box knowledge distillation leverages the output distributions and hidden states of teacher models to provide the student model with richer learning signals \cite{gu2024minillm, team2024gemma, kodistillm}. However, this approach is infeasible for black-box settings where access to internal teacher model states is restricted. The Chain-of-Thought paradigm has further advanced knowledge distillation by enabling the generation of detailed reasoning samples from teacher models \cite{ho2022largelanguagemodels}. These samples allow student models to learn both the correct answers and the intricate reasoning processes behind them. \cite{hsieh-etal-2023-distilling}. To enhance consistency, recent efforts have focused on generating diverse rationales for the same query \cite{chen2023mcc, liu2023mind}, ensuring more robust predictions. However, relying on rationales from a single teacher model risks introducing bias and limiting the scope of reasoning diversity. Multi-teacher learning paradigms, which integrate knowledge from multiple teacher models, hold the potential to overcome these limitations by enhancing knowledge diversity and capturing a broader range of reasoning styles. 

\section{Method}
In this section, we formally present \ours to resolve the challenges described in the Introduction. In particular, we start by describing the preliminary. Next, we introduce the details of \ours by first obtaining rationales from multiple teachers, and then learning a small student using the obtained rationales. The \ours pipeline is shown in Figure ~\ref{fig:pipeline}.

\subsection{Preliminary}
\textbf{Multiple Choice Question Answering.} A $k$-way multiple choice question answering (MCQA) is defined as follows: Given a question $Q_i$, a set of candidate answer options $O_i=\{O_{i1},O_{i2},...,O_{ik}\}$, the model is tasked with selecting the correct answer from the set $O_i$, such that the selected answer aligns the ground truth label $A_i$.

\noindent \textbf{Knowledge Distillation.} 
The knowledge distillation process begins with the teacher model, denoted as $T$ parameterized by $\theta_T$, which has been pre-trained on a large corpus.
Later, the student model, $S$, with parameter $\theta_S$, is tasked with distilling knowledge directly from $T$, leveraging the strong capabilities of $T$. Correspondingly, the objective function can be formulated as: $\mathcal{L} = \ell(S, T)$, where $\ell$  indicates the learning function, e.g., cross-entropy loss between the prediction output of the student and the target output generated by the teacher.

\subsection{Obtaining Rationales from Teachers}

To enforce the teacher LLMs generate high-quality rationales for student learning, our approach incorporates an In-context Example Generator to enrich contextual information, a Teacher-forcing Chain-of-Thought strategy to avoid misinformation, and rationale generation from multiple teachers to support student training.

\noindent
\textbf{In-context Example Generator.}
To enable that rationales generated by teacher models are grounded in contextually appropriate scenarios, we introduce an optional in-context example generator. This tool provides additional context by generating examples that include both questions and corresponding rationales in a zero-shot setting, enhancing the teacher models' comprehension of task-specific nuances. Each generated in-context example offers a unique perspective on the dataset and provides additional insights. By including a range of in-context examples and sending them with the input question to the teacher LLMs, we enhance the input with richer information that aids the model in generating higher-quality rationales. This approach allows the student model to learn not only from correct answers but also from the underlying reasoning, thereby enhancing both the accuracy and interpretability of the distilled model.

\noindent
\textbf{Teacher-forcing Chain-of-Thought.}
In addition, we design a teacher-forcing strategy to ensure the validity of the rationales. Compared to existing methods that simply employ regular chain-of-thought (CoT) mechanisms \cite{wei2022chainofthought,kojima2022zero}, wherein an LLM is prompted with sets of questions and options \(\{Q_i, O_i\}\) to elicit rationales \(R_i\) directly, \ours posits a distinct advantage in integrating the correct answer \(A_i\) into the input. We hypothesize that the inclusion of \(A_i\) alongside \(Q_i\) and \(O_i\) facilitates a more nuanced understanding of the input context and the correct logical rationales leading to the answer, thereby facilitating a more informed and accurate generation process. 
Specifically, we consider the concatenation of questions, options, and answers \(\{Q_i, O_i, A_i\}\) as the input to LLMs.

\noindent
\textbf{Rationales from Multiple Teachers.}
Given $M$ teachers, \ours pioneers the usage of a multi-teacher architecture in which each teacher \(T^m\) is an LLM.
In particular, the rationale \(R_i^m\) produced by a specific teacher model $\theta_{T^m}$ for the \(i\)th question is derived using the question $Q_i$, options $O_i$, correct answer $A_i$, and in-context examples $P_i$.
The process is formalized as follows:
\begin{equation}
R_i^m = T^m(Q_i, O_i, A_i, P_i; \theta_{T^m}).
\end{equation}
\subsection{Learning a Small Student}
A straightforward strategy to incorporate rationales as supervision is to append each rationale \(R_i^m\) generated by the teacher models as supplementary input to the student model, along with the question \(Q_i\) and options \(O_i\). 
However, this method faces challenges due to limitations in computational resources at the inference stage, especially because rationales must be pre-generated for every data sample in both training and test sets \cite{wang2022pinto}. 
To overcome this issue, we employ rationales as a form of supervisory signal during the training process to develop a model that is adept at generating its explanations. Subsequently, this trained model can be utilized on the test set, eliminating the need for pre-generated rationales to facilitate accurate reasoning.
Specifically,
\ours integrates rationales from multiple teacher models into a unified multi-task instruction tuning framework. This necessitates the assignment of a unique prefix \(p\) for distinguishing between learning tasks from different teachers. The student model is trained not only to predict labels but also to generate rationales akin to those produced by the teachers. 
Accordingly, the overall loss function $\mathcal{L}$ is as follows:
\begin{equation}
\mathcal{L} = \mathcal{L}_A + \sum_{m=1}^{M}\alpha^m\mathcal{L}_{T^m}, 
\label{eq:total_loss}
\end{equation}
where \(\mathcal{L}_A\) denotes the objective of learning from ground truth answers, $\mathcal{L}_{T^m}$ indicates the objective of learning from $m$-th teacher, 
$\alpha^m$ is the importance weight for $T^m$, and $M$ is the number of teacher LLMs. 
Formally, $\mathcal{L}_A$ and $\mathcal{L}_{T^m}$ are defined as follows:
\begin{equation}
\mathcal{L}_A = \frac{1}{N} \sum_{i=1}^{N} \ell(S(Q_i, O_i, p_A; \theta_S), A_i),
\end{equation}
\begin{equation}
    \mathcal{L}_{T^m} = \frac{1}{N} \sum_{i=1}^{N} \ell(S(Q_i, O_i, p_m; \theta_S), R_i^m),
\end{equation}
where $N$ is the number of data samples, $\ell$ indicates the cross-entropy loss between the predicted and target tokens. Here $\mathcal{L}_A$ encourages the student $S$ to generate ground truth answer $A_i$ by minimizing the difference between it and the student output given the question $Q_i$, options $O_i$, and instruction prefix $p_A$ for generating answers. On the other hand, $\mathcal{L}_T^m$ facilitates the student $S$ to mimic the reasoning capability of teacher $T^m$ by learning from its rationale $R_i^m$, with the guidance of instruction prefix $p_m$ for $T^m$.

\begin{table*}[t]
\centering
\caption{Overall experimental results. The best results across different datasets and LLM sizes are highlighted in bold. $\Delta_{FF}$ and $\Delta_{Distill}$ represent the relative performance improvement of \ours to Full Fine-Tuning and Distill-step-by-step, respectively. Accuracy is used as the evaluation metric.}
  \begin{tabular}{ccccccccc}
    \toprule
     & & \multicolumn{4}{c}{Commonsense Reasoning} & \multicolumn{2}{c}{Biomedical Reasoning} & \\ 
 \cmidrule{3-8}
     \multirow{-2.3}{*}{\textbf{Setting}} & \multirow{-2.3}{*}{\textbf{Method}} & {\textbf{OBQA}} & {\textbf{ARC}} & {\textbf{PIQA}} & {\textbf{Riddle}} & {\textbf{PQA}} & {\textbf{BioASQ}} & \multirow{-2.3}{*}{\textbf{Total}}\\
    \midrule
     \multirow{-0.5}{*}{3B/7B Teacher} & FLAN-T5 xlarge & 69.20 & 68.24 & 58.43 & 53.73 & 71.50 & 65.85 & 64.49\\
     & LLaMA 2 & 58.60 & 45.90 & 78.80 & 47.65 & 54.50 & 73.75 & 59.87\\
    \midrule
     \multirow{9}{*}{\shortstack{80M Student\\Size: 2.7\%/1.1\%}} & Inference & 16.60 & 19.31 & 20.78 & 13.33 & 38.00 & 47.97 & 26.00 \\
     & {PINTO} & 46.40 & 26.87 & 48.10 & 25.29 & 60.00 & 80.49 & 47.86 \\
     & LoRA & 37.80 & 27.12 & 39.93 & 39.80 & 53.75 & 78.05 & 46.08 \\
     & {Full Fine-tuning} & 41.60 & 27.47 & 42.33 & 42.75 & 56.25 & 78.86 & 48.21\\
     & {Standard KD} & 45.80 & 29.53 & 49.29 & 36.27 & 58.00 & 81.30 & 49.43\\
     & {Distill-step-by-step} & 46.40 & 30.47 & 50.38 & 36.67 & 59.00 & 81.30 & 50.70\\
     & {\ours} & \textbf{49.40} & \textbf{33.05} & \textbf{53.65} & \textbf{51.18} & \textbf{62.00} & \textbf{85.37} & \textbf{55.78}\\
     & {$\Delta_{FF}$} & $\uparrow 18.75\%$ & $\uparrow 20.31\%$ & $\uparrow 26.74\%$ & $\uparrow 19.72\%$ & $\uparrow 10.22\%$ & $\uparrow 8.26\%$ & $\uparrow 15.69\%$\\
     & {$\Delta_{Distill}$} & $\uparrow 6.47\%$ & $\uparrow 8.47\%$ & $\uparrow 6.49\%$ & $\uparrow 39.57\%$ & $\uparrow 5.08\%$ & $\uparrow 5.01\%$ & $\uparrow 10.00\%$\\
     
    \midrule
    \multirow{9}{*}{\shortstack{250M Student\\Size: 8.3\%/3.6\%}} & Inference & 31.00 & 23.00 & 30.47 & 30.78 & 48.00 & 57.72 & 36.83 \\
    & {PINTO} & 50.40 & 38.63 & 52.12 & 34.90 & 61.75 & 82.93 & 53.46 \\
    & LoRA & 51.40 & 37.25 & 47.66 & 53.14 & 62.00 & 82.93 & 55.73\\
    & {Full Fine-tuning} & 56.60 & 38.88 & 47.55 & 52.55 & 64.75 & 89.43 & 58.29\\
    & {Standard KD} & 55.40 & 43.69 & 55.93 & 42.94 & 64.25 & 86.18 & 58.07 \\
    & {Distill-step-by-step} & 56.80 & 43.86 & 56.37 & 45.69 & 64.75 & 86.18 & 58.94 \\
    & {\ours} & \textbf{64.20} & \textbf{48.50} & \textbf{60.17} & \textbf{60.78} & \textbf{66.25} & \textbf{90.24} & \textbf{65.02}\\
    & {$\Delta_{FF}$} & $\uparrow 13.43\%$ & $\uparrow 24.74\%$ & $\uparrow 26.54\%$ & $\uparrow 15.66\%$ & $\uparrow 2.32\%$ & $\uparrow 0.91\%$ & $\uparrow 11.55\%$\\
    & {$\Delta_{Distill}$} & $\uparrow 13.03\%$ & $\uparrow 10.58\%$ & $\uparrow 6.74\%$ & $\uparrow 33.03\%$ & $\uparrow 2.32\%$ & $\uparrow 4.71\%$ & $\uparrow 10.32\%$ \\
    
    \midrule
    \multirow{9}{*}{\shortstack{780M Student\\Size: 26.0\%/11.1\%}} & Inference & 50.40 & 51.07 & 51.90 & 39.80 & 64.25 & 63.41 & 53.47\\
    & {PINTO} & 62.20 & 52.10 & 57.13 & 42.94 & 70.00 & 84.55 & 61.49 \\
    & LoRA & 64.00 & 57.77 & 57.02 & 68.63 & 70.25 & 86.18 & 67.31\\
    & {Full Fine-tuning} & 71.20 & 62.92 & 58.43 & 68.82 & 70.25 & 90.24 & 70.31\\
    & {Standard KD} & 65.80 & 56.05 & 60.72 & 52.94 & 70.00 & 86.99 & 65.42 \\
    & {Distill-step-by-step} & 66.80 & 57.42 & 61.37 & 53.92 & 70.00 & 86.99 & 66.08 \\
    & {\ours} & \textbf{74.40} & \textbf{64.29} & \textbf{67.90} & \textbf{70.98} & \textbf{73.00} & \textbf{92.68} & \textbf{73.88}\\
    & {$\Delta_{FF}$} & $\uparrow 4.49\%$ & $\uparrow 2.18\%$ & $\uparrow 16.21\%$ & $\uparrow 3.14\%$ & $\uparrow 3.91\%$ & $\uparrow 2.70\%$ & $\uparrow 5.07\%$\\
    & {$\Delta_{Distill}$} & $\uparrow 11.38\%$ & $\uparrow 11.96\%$ & $\uparrow 10.64\%$ & $\uparrow 31.64\%$ & $\uparrow 4.29\%$ & $\uparrow 6.54\%$ & $\uparrow 11.79\%$\\
    
    \bottomrule
  \end{tabular}

\label{Overall}
\end{table*}

\section{Experiments}
In this section, we rigorously test \ours against a series of empirical benchmarks across varied datasets and reasoning tasks. 
In addition, we conduct efficiency analyses, ablation studies, parameter sensitivity tests, and case studies to demonstrate the effectiveness and superiority of our method.

\subsection{Experimental Setup}

\textbf{Datasets.} 
For the task of commonsense reasoning, 
we use OpenBookQA (OBQA) \cite{OpenBookQA2018}, The AI2 Reasoning Challenge (ARC) \cite{Clark2018ThinkYH}, Physical Interaction Question Answering (PIQA) \cite{Bisk2020PIQA}, and RiddleSense (Riddle) \cite{lin-etal-2021-riddlesense}. For the task of biomedical reasoning, we consider PubMedQA (PQA) \cite{jin2019pubmedqa} and BioASQ \cite{Tsatsaronis2015BIOASQ}.

\noindent \textbf{Baselines.} 
We benchmark \ours against the teacher’s performance and various baseline methods, including Inference-only that only leverage the pre-trained model for evaluation without  training, and multiple fine-tuning methods that provide further adaptation. In particular, we consider LoRA \cite{hu2022lora}, full fine-tuning, and the PINTO method \cite{wang2022pinto} for the fine-tuning methods.
We also compare \ours with various knowledge distillation strategies. To illustrate, we include standard KD \cite{hinton_kd} that enforces the student to mimic the teacher's labels and the Distill-step-by-step method \cite{hsieh-etal-2023-distilling} that leverage rationales. 

\noindent \textbf{Implementation Details.}  
For all distillation baselines and \ours, we set the learning rate to \(5 \times 10^{-5}\), batch size to 8, maximum input length to 1024, and epoch to 1. 
For Distill-step-by-step and \ours, the trade-off weights \(\alpha_{T_n}\) are explored within \{0.01, 0.1, 0.5, 1, 2, 3\}.
We report the best result for Distill-step-by-step by leveraging different teacher models.
For the choice of LLMs, we use FLAN-T5 \cite{chung2022scaling} small (80M), base (250M), and large (780M) as the student, and FLAN-T5 xlarge (3B) and LLaMA 2-chat \cite{touvron2023llama2} (7B) as teachers. Experiments are conducted on four NVIDIA H100 Tensor Core GPUs.
\subsection{Performance Comparison}

\noindent \textbf{Comparison to Baselines Methods.}
We conducted a thorough evaluation of our method, \ours, across six diverse datasets spanning two distinct reasoning tasks: commonsense reasoning and biomedical reasoning. The detailed results are presented in Table ~\ref{Overall}, offering a comprehensive view of the performance landscape across different model sizes and methodologies. From the table, we derive several observations as follows.

First, while full fine-tuning is theoretically capable of maximizing parameter adjustments and should yield the best results in principle, it does not consistently outperform parameter-efficient fine-tuning methods such as LoRA. This outcome suggests that simply having a larger number of adjustable parameters does not guarantee improved generalization or performance, especially when the fine-tuning process may inadvertently overfit the training data or fail to capture nuanced task-specific knowledge.

Second, \ours demonstrates consistent and substantial performance improvements across all datasets and model sizes, underscoring the robustness and adaptability of our approach. Specifically, the quantitative gains achieved by \ours are noteworthy: we observe an average performance boost of \textbf{+15.69\%}, \textbf{+11.55\%}, and \textbf{+5.07\%} for student models with 80M, 250M, and 780M parameters, respectively, compared to full fine-tuning. These improvements are significant, particularly considering that larger models often exhibit diminishing returns on performance gains, making these results even more compelling. 

Third, when compared to the state-of-the-art distillation method, Distill-step-by-step, \ours achieves impressive relative improvements of \textbf{+10.00\%}, \textbf{+10.32\%}, and \textbf{+11.79\%} for the 80M, 250M, and 780M student models, respectively. These results highlight the effectiveness of our method in distilling knowledge from large models into the smaller student model. The consistent superiority of \ours across different model sizes and datasets demonstrates the advantages of our design choices, including the strategic balance of integrating intermediate rationale guidance and mimicking teacher outputs. Furthermore, using rationales during training enhances the model's reasoning capability and improves interpretability over conventional distillation.

\begin{figure*}[htbp]
\begin{center}
\includegraphics[width=0.95\textwidth]{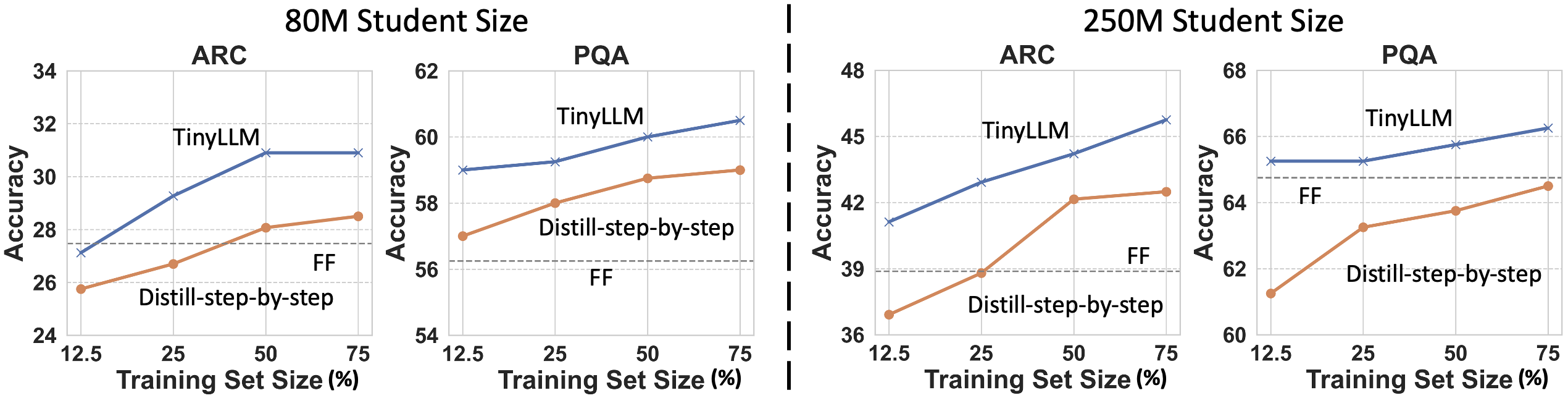}
\end{center}
\caption{
A comparative analysis of \ours against the state-of-the-art Distill-step-by-step method using 80M and 250M FLAN-T5 model architectures across various training set sizes. Dotted line indicates the full fine-tuning (FF) using 100\% dataset.
It is evident that \ours consistently surpasses the performance of both Distill-step-by-step and full fine-tuning. Notably, \ours achieves this superior accuracy while employing substantially fewer training examples. 
}
\label{fig:reduction}
\end{figure*}

\noindent \textbf{Comparison to Teachers.}
In addition to outperforming baseline methods, \ours also exhibits superior performance compared to the original teacher models. This is particularly remarkable given the significant difference in model size. For instance, a 780M parameter student model trained with \ours achieves an impressive average performance score of 73.88 across various datasets. This score represents a substantial improvement of \textbf{+14.56\%} over the performance of 3B teacher model, and an even more striking \textbf{+23.40\%} performance improvement over the 7B teacher model. However, further reducing the size of the student model can introduce difficulties for the student to learn and result in decreased performance.

These results demonstrate the efficacy of \ours in transferring knowledge and suggest that our approach enables smaller models to generalize better and perform tasks more effectively than their larger counterparts. Furthermore, the efficiency gains are particularly pronounced when considering smaller student models, such as the 250M parameter model, which manages to surpass both the 3B and 7B parameter teachers with relative improvements of \textbf{+0.82\%} and \textbf{+8.60\%}, respectively. It is also worth mentioning that the 250M model operates with only \textbf{8.3\%} and \textbf{3.6\%} of the teacher models' parameters, showcasing the remarkable efficiency of \ours in compressing and enhancing the performance of smaller models without compromising accuracy.

\subsection{Efficiency Analysis of Training Set Size in Knowledge Transfer} 

\noindent
\textbf{Advantage Over Standard Knowledge Distillation.}
To thoroughly assess the efficiency and effectiveness of our proposed method, \ours, we conducted a series of experiments that evaluate its performance across varying training set sizes, particularly in comparison to the state-of-the-art Distill-step-by-step method. This analysis is crucial for understanding how well our model performs not only with full datasets but also under conditions of limited training data, which is often a real-world constraint.

As illustrated in Figure ~\ref{fig:reduction}, \ours consistently outperforms the Distill-step-by-step method across all tested ratios of the training data, demonstrating its robustness and efficiency. This superior performance is particularly evident even when we reduce the size of the training set. The trend indicates that \ours can effectively leverage smaller amounts of data to achieve comparable or better results than methods that require larger training datasets.

A striking example of this efficiency is observed in the context of the PQA dataset. Here, \ours, when trained with only 12.5\% of the available training data, not only meets but often exceeds the performance levels achieved by Distill-step-by-step when utilizing the full training set. This significant reduction in required training data, without sacrificing performance, underscores the practical advantages of \ours, particularly in scenarios where data is scarce or expensive to obtain.

This finding holds across different model sizes, including both 80M and 250M parameter student models. The consistent outstanding performance of \ours in these cases suggests that our method is highly effective in transferring knowledge efficiently, making it a versatile tool for various applications, regardless of the model size. The capability of achieving superior performance with limited data can be particularly beneficial in resource-constrained environments, where computational resources and training time are restricted.

\noindent \textbf{Outperforming Full Fine-Tuning.}
Beyond its advantages over standard knowledge distillation methods, \ours also demonstrates significant improvements compared to traditional full fine-tuning approaches, even when using the entire dataset. This comparison further highlights the efficiency of our method in terms of both data usage and computational resources.

In particular, when training a 250M parameter model with \ours on the ARC and PQA datasets, as well as training an 80M parameter model on the PQA dataset, only 12.5\% of the full training data is necessary to surpass the benchmarks established by full fine-tuning. This result is remarkable, as it shows that \ours can achieve superior performance with a fraction of the data typically required for full fine-tuning, significantly reducing the computational cost and time required for training.

In addition, for 80M student model trained on the ARC dataset, \ours can achieve a higher accuracy than full fine-tuning with a 75\% reduction in training samples, while further increasing the reduction to 87.5\% results in a competitive performance. This result demonstrates the capability of \ours in reducing the need for extensive training data to maintain or further improve the model performance, making it an attractive alternative to full fine-tuning. 

\begin{table}[t]
  \centering
  \caption{Impact of in-context examples, the teacher-forcing strategy and contributions of various teachers.}
  \resizebox{\linewidth}{!}{
  \begin{tabular}{ccccccc}
    \toprule
    &  \multicolumn{4}{c}{Commonsense} & \multicolumn{2}{c}{Biomedical} \\ 
    \cmidrule{2-7}
    \multirow{-2.3}{*}{\textbf{Variant}} & {\textbf{OBQA}} & {\textbf{ARC}} & {\textbf{PIQA}} & {\textbf{Riddle}} & {\textbf{PQA}} & {\textbf{BioASQ}} \\
    \midrule
    {w/o in-context}  & 73.20 & 63.09 & 66.27 & 69.22 & 70.75 & 86.99 \\
    
    {w/o LLaMA}  & 73.00 & 62.32 & 66.70 & 68.82 & 69.25 & 87.81 \\

    {w/o T5}  & 73.80 & 61.80 & 66.49 & 68.63 & 69.50 & 88.62 \\
    
    {w/o diverse teachers}  & 73.80 & 62.49 & 66.81 & 68.82 & 70.00 & 89.43 \\
    
    {w/o teacher-forcing}  & 73.80 & 60.94 & 65.94 & 69.02 & 70.25 & 90.24 \\

    {\ours} & \textbf{74.40} & \textbf{64.29} & \textbf{67.90} & \textbf{70.98} & \textbf{73.00} & \textbf{92.68} \\
    \bottomrule
  \end{tabular}
  }

\label{tab:ablation_study}
\end{table}

\subsection{Ablation Study}
To provide a comprehensive evaluation of our proposed method, \ours, we conducted an ablation study in Table~\ref{tab:ablation_study} to validate the contributions of key components in enhancing the reasoning capabilities of the distilled LLM. Specifically, we focused on assessing the impact of the in-context example generator, the use of rationales from multiple teacher models, and the teacher-forcing strategy. By isolating these components, we aim to understand their individual and collective contributions to the overall performance of \ours.
We designed four ablation variants of \ours, each purposefully modified to test the significance of specific components:

\begin{itemize}[nosep,leftmargin=*]
\item \textbf{w/o in-context} removes the use of in-context examples during rationale generation. This variant is crucial for evaluating the role of in-context examples in guiding the student model to produce more accurate, relevant, and contextually appropriate rationales. 

\item \textbf{w/o LLaMA} and \textbf{w/o T5} exclude the rationale supervision provided by the respective teacher models during the distillation process. By removing the influence of either LLaMA or T5, these variants help us understand the individual contributions of each teacher's rationales. 

\item \textbf{w/o diverse teachers} excludes the weaker teacher model, instead generating multiple rationales from the stronger teacher model. This variant is designed to test the effectiveness of using a diverse set of teachers, as opposed to relying on a single, potentially more robust teacher. 

\item \textbf{w/o teacher-forcing} eliminates the teacher-forcing strategy during rationale generation. Teacher-forcing is a technique where the model is trained using the correct output (from the teacher) as input for the next step, rather than its own previous prediction. By removing this strategy, we aim to assess its effectiveness in helping the student model generate higher-quality rationales. 

\end{itemize}

Table~\ref{tab:ablation_study} provides a comparative analysis between each ablation variant and the complete \ours model. From the data presented in Table~\ref{tab:ablation_study}, several important observations emerge: (1) \ours consistently outperforms all ablation variants, demonstrating that the integration of all components—rationales from multiple teachers, in-context examples, and the teacher-forcing strategy—collectively contributes to the model's superior performance. This result highlights the synergistic effect of these components, where their combination leads to significant improvements in the reasoning capabilities of the distilled LLM. (2) There is no substantial performance gap between the different ablation variants, suggesting that while each component contributes to the model's performance, none of them alone is solely responsible for the observed improvements. This finding implies a balanced importance among the components, with each playing a complementary role in refining the model's reasoning abilities.

\subsection{Parameter Sensitivity}

\begin{figure}
  \centering
  \includegraphics[width=\linewidth]{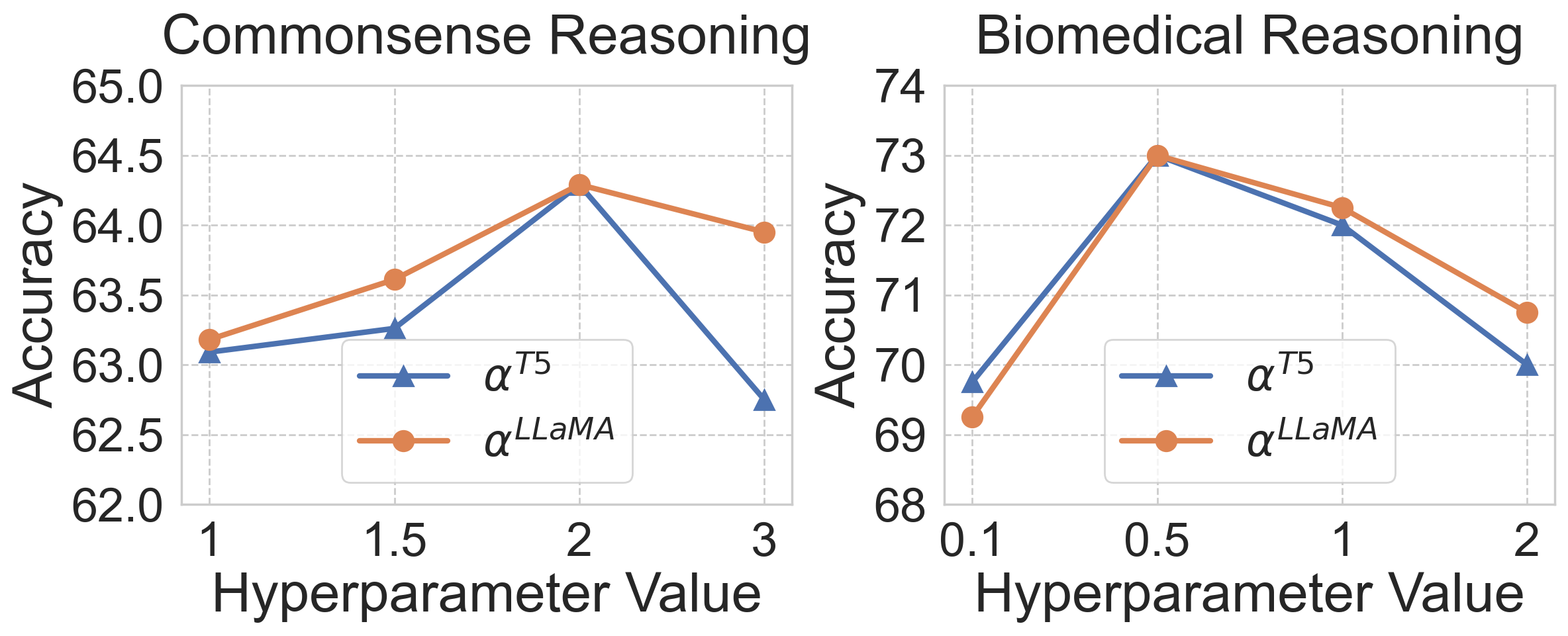}
  \caption{Performance w.r.t. different values of weight $\alpha$.}
  \label{fig:sensitivity}
\end{figure}

To thoroughly evaluate the robustness and adaptability of our proposed model, we conducted parameter sensitivity experiments on two datasets in different tasks: ARC for commonsense reasoning and PQA for biomedical reasoning. These experiments are critical for understanding how different parameter settings, particularly the trade-off weights $\alpha^{T5}$ and $\alpha^{LLaMA}$, influence the model's performance across various tasks. The results of these sensitivity analyses are depicted in Figure ~\ref{fig:sensitivity}.

In this analysis, our primary focus is on exploring the effects of varying the trade-off weights $\alpha^{T5}$ and $\alpha^{LLaMA}$, which balance the influence of rationales provided by the T5 and LLaMA teacher models, respectively. This exploration reveals the model's adaptability and how it responds to different parameter configurations, offering insights into the optimal settings for maximizing performance across different tasks and datasets.
From the results shown in Figure ~\ref{fig:sensitivity}, our key observations are as follows.

\noindent
\textbf{Optimal Parameter Variability Across Datasets and Tasks.}
It is evident that the optimal parameter settings for $\alpha^{T5}$ and $\alpha^{LLaMA}$ vary depending on the dataset and the nature of the reasoning task. For biomedical reasoning tasks, such as those found in the PQA dataset, questions tend to be lengthy and complex, often requiring deep comprehension and retrieval of detailed information. In these cases, the impact of rationales from the teacher models is partially diminished, given the complexity of the content overshadows the direct utility of rationale guidance. As a result, a smaller value of $\alpha$ is sufficient to achieve optimal performance. On the other hand, commonsense reasoning tasks, exemplified by the ARC dataset, typically involve more concise and straightforward questions. In these scenarios, the rationales provided by the teacher models play a more critical role in guiding the student model's reasoning process, leading to the need for a larger $\alpha$ value to fully leverage this guidance.

\noindent
\textbf{Effect of Increasing $\alpha$ Values on Performance.}
Increasing the values of $\alpha^{T5}$ and $\alpha^{LLaMA}$ generally leads to improved performance, which can be attributed to the enhanced Chain-of-Thought reasoning capabilities of the student model. By placing greater emphasis on the rationales from teacher models, the student model benefits from a richer multi-task learning experience, which enhances its prediction capabilities. However, this improvement has its limits. Excessively high values of $\alpha$ can degrade performance by causing the model to overly focus on reasoning processes at the expense of prediction accuracy. This shift in focus may lead to a situation where the model becomes too reliant on rationales, potentially overfitting to the rationale structure rather than learning to generalize effectively from the data.

\noindent
\textbf{Differential Sensitivity of $\alpha^{T5}$ and $\alpha^{LLaMA}$ Across Datasets.}
The sensitivity of the parameters $\alpha^{T5}$ and $\alpha^{LLaMA}$ varies between different datasets, reflecting the diverse contributions of the T5 and LLaMA teacher models to the reasoning process. For commonsense reasoning tasks, we observe that $\alpha^{T5}$ is more sensitive than $\alpha^{LLaMA}$. This suggests that the rationales generated by the T5 model are particularly valuable for tasks involving straightforward reasoning, where the logical flow and concise reasoning provided by T5 are more impactful. Conversely, in biomedical reasoning tasks, the sensitivities of $\alpha^{T5}$ and $\alpha^{LLaMA}$ are more balanced, indicating that both teacher models offer valuable, albeit different, insights that complement each other in processing the complex and detailed content typical of biomedical texts. This balance highlights the importance of leveraging diverse teacher models to capture the full range of reasoning required for different tasks.

\subsection{Case Study}

\begin{figure*}[htbp!]
\begin{center}
\includegraphics[width=\textwidth]{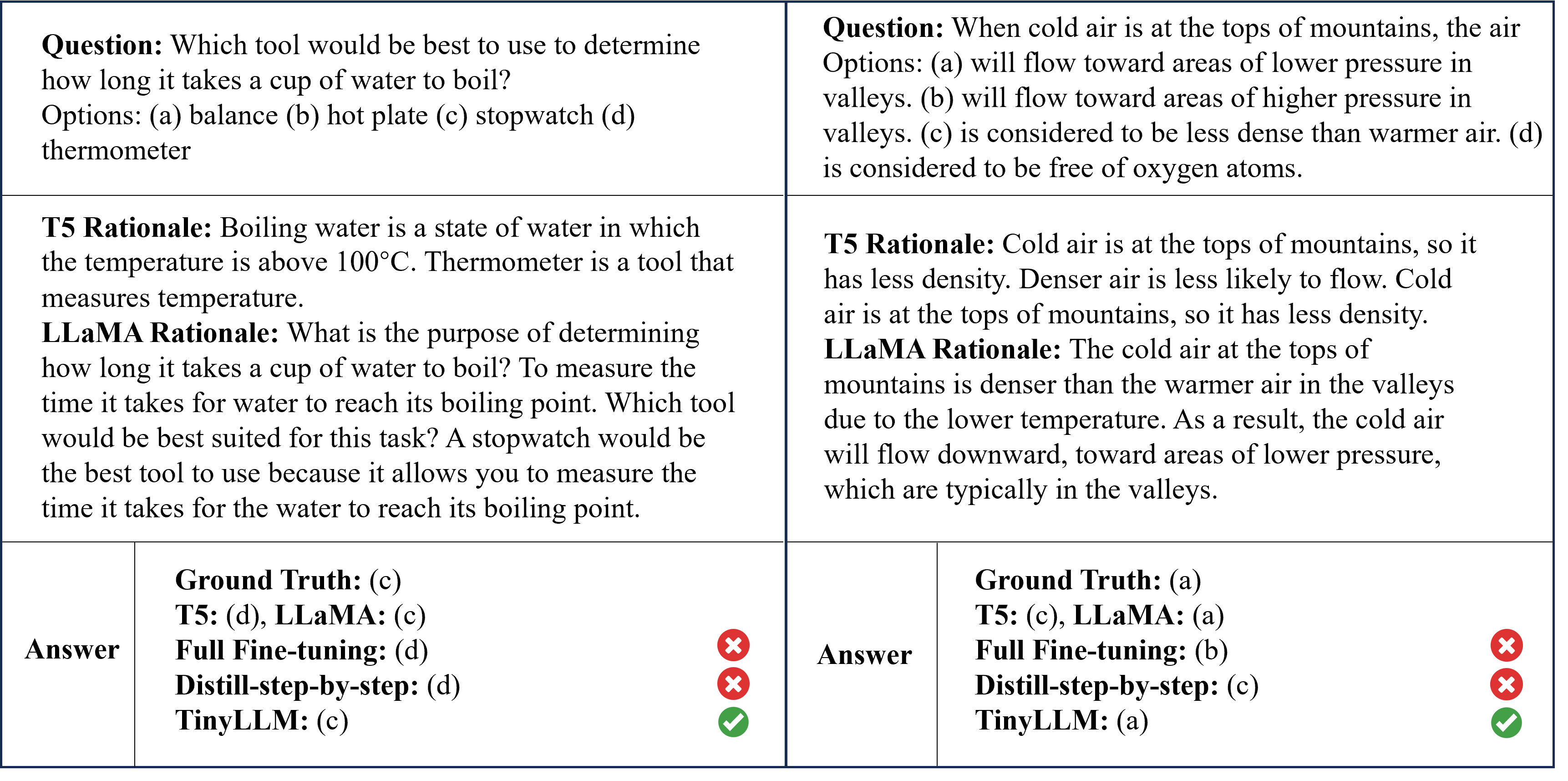}
\end{center}
\caption{
Case study of different models' prediction.
Examples are selected from the ARC and PIQA datasets. In both cases, \ours successfully generates the correct answer. 
}
\label{fig:case_study}
\end{figure*}

To gain a more intuitive understanding of why \ours consistently outperforms other models, we conducted case studies by comparing the predictions generated by different models. These case studies offer valuable insights into the specific scenarios where \ours excels and highlight the advantages of our multi-teacher approach. Figure ~\ref{fig:case_study} presents two representative examples that we randomly selected from the ARC and PIQA datasets, illustrating the differences in model predictions and the underlying rationales.

In the first example, taken from the ARC dataset, we observe a significant discrepancy in the reasoning capabilities of the teacher models. The T5 model provides a completely incorrect rationale, leading to an incorrect prediction. In contrast, LLaMA generates a meaningful and accurate rationale, which correctly guides the prediction. However, the full fine-tuning method, which does not utilize teacher rationales, also fails to provide the correct answer. Similarly, the state-of-the-art Distill-step-by-step method also predicts incorrectly, likely because the distillation process, which leverages T5's reasoning, introduces noise that misguides the student model. This occurs even though T5 significantly outperforms LLaMA in overall accuracy on the ARC dataset (as shown in Table \ref{Overall}). However, \ours demonstrates its robustness by correctly inferring the answer (a), effectively understanding the rationales from both T5 and LLaMA, and filtering out the noise introduced by the incorrect rationale from T5. This example highlights the strength of \ours in balancing and integrating multiple sources of knowledge, leading to superior prediction accuracy.

A similar situation occurs in the second example from the PIQA dataset. T5 provides an incorrect rationale, while LLaMA aligns with the ground truth. Other methods failed to predict the correct answer. However, \ours successfully integrates and balances rationales from both teachers. 

\section{Conclusion and Future Work}
In this paper, we propose \ours, a novel knowledge distillation paradigm to learn a small student LLM from multiple large teacher LLMs.
\ours involves several principled designs, such as learning contextually appropriate rationales using an in-context example generator, enabling the credibility of rationales with a teacher-forcing Chain-of-Thought strategy, and inheriting a wider range of knowledge from various teachers.
Our extensive empirical evaluation and in-depth analysis, conducted across six datasets spanning two reasoning tasks, demonstrate that \ours brings significant and consistent improvements by up to \textbf{+15.69\%} over full fine-tuning, up to \textbf{+23.40\%} over teacher models, and up to \textbf{+11.79\%} over state-of-the-art.
Moreover, \ours holds a significantly smaller model size, e.g., \textbf{1.1\%} to \textbf{26.0\%} compared to the sizes of the teachers.

As future work, we envision several promising directions to further enhance \ours. One potential avenue is to manage any conflicting rationales that can arise when integrating a broader set of teacher LLMs. Conflicting rationales can emerge due to hallucinations, variations in reasoning styles, or domains of expertise for different teacher models.
Another direction involves using open-source embeddings to represent questions within each dataset. This approach maps questions into a semantic space and allows us to select in-context examples based on similarity or the top nearest neighbors.
By incorporating targeted in-context example selection, we can assess whether providing such focused and meaningful context can enhance the comprehension and performance of the student model.

\balance
\bibliographystyle{ACM-Reference-Format}
\bibliography{sample-base}


\begin{thebibliography}{84}


\ifx \showCODEN    \undefined \def \showCODEN     #1{\unskip}     \fi
\ifx \showDOI      \undefined \def \showDOI       #1{#1}\fi
\ifx \showISBNx    \undefined \def \showISBNx     #1{\unskip}     \fi
\ifx \showISBNxiii \undefined \def \showISBNxiii  #1{\unskip}     \fi
\ifx \showISSN     \undefined \def \showISSN      #1{\unskip}     \fi
\ifx \showLCCN     \undefined \def \showLCCN      #1{\unskip}     \fi
\ifx \shownote     \undefined \def \shownote      #1{#1}          \fi
\ifx \showarticletitle \undefined \def \showarticletitle #1{#1}   \fi
\ifx \showURL      \undefined \def \showURL       {\relax}        \fi
\providecommand\bibfield[2]{#2}
\providecommand\bibinfo[2]{#2}
\providecommand\natexlab[1]{#1}
\providecommand\showeprint[2][]{arXiv:#2}

\bibitem[Agarwal et~al\mbox{.}(2024)]%
        {agarwal2024generalized}
\bibfield{author}{\bibinfo{person}{Rishabh Agarwal}, \bibinfo{person}{Nino Vieillard}, \bibinfo{person}{Yongchao Zhou}, \bibinfo{person}{Piotr Stanczyk}, \bibinfo{person}{Sabela~Ramos Garea}, \bibinfo{person}{Matthieu Geist}, {and} \bibinfo{person}{Olivier Bachem}.} \bibinfo{year}{2024}\natexlab{}.
\newblock \showarticletitle{Generalized Knowledge Distillation for Auto-regressive Language Models}. In \bibinfo{booktitle}{\emph{The Twelfth International Conference on Learning Representations}}.
\newblock


\bibitem[Anil et~al\mbox{.}(2023)]%
        {anil2023palm}
\bibfield{author}{\bibinfo{person}{Rohan Anil}, \bibinfo{person}{Andrew~M Dai}, \bibinfo{person}{Orhan Firat}, \bibinfo{person}{Melvin Johnson}, \bibinfo{person}{Dmitry Lepikhin}, \bibinfo{person}{Alexandre Passos}, \bibinfo{person}{Siamak Shakeri}, \bibinfo{person}{Emanuel Taropa}, \bibinfo{person}{Paige Bailey}, \bibinfo{person}{Zhifeng Chen}, {et~al\mbox{.}}} \bibinfo{year}{2023}\natexlab{}.
\newblock \showarticletitle{Palm 2 technical report}.
\newblock \bibinfo{journal}{\emph{arXiv preprint arXiv:2305.10403}} (\bibinfo{year}{2023}).
\newblock


\bibitem[Bang et~al\mbox{.}(2023)]%
        {bang2023multitaskmultilingual}
\bibfield{author}{\bibinfo{person}{Yejin Bang}, \bibinfo{person}{Samuel Cahyawijaya}, \bibinfo{person}{Nayeon Lee}, \bibinfo{person}{Wenliang Dai}, \bibinfo{person}{Dan Su}, \bibinfo{person}{Bryan Wilie}, \bibinfo{person}{Holy Lovenia}, \bibinfo{person}{Ziwei Ji}, \bibinfo{person}{Tiezheng Yu}, \bibinfo{person}{Willy Chung}, \bibinfo{person}{Quyet~V. Do}, \bibinfo{person}{Yan Xu}, {and} \bibinfo{person}{Pascale Fung}.} \bibinfo{year}{2023}\natexlab{}.
\newblock \showarticletitle{A Multitask, Multilingual, Multimodal Evaluation of ChatGPT on Reasoning, Hallucination, and Interactivity}. In \bibinfo{booktitle}{\emph{ACL}}.
\newblock


\bibitem[Bisk et~al\mbox{.}(2020)]%
        {Bisk2020PIQA}
\bibfield{author}{\bibinfo{person}{Yonatan Bisk}, \bibinfo{person}{Rowan Zellers}, \bibinfo{person}{Ronan Le~Bras}, \bibinfo{person}{Jianfeng Gao}, {and} \bibinfo{person}{Yejin Choi}.} \bibinfo{year}{2020}\natexlab{}.
\newblock \showarticletitle{PIQA: Reasoning about Physical Commonsense in Natural Language}. In \bibinfo{booktitle}{\emph{AAAI}}.
\newblock


\bibitem[Brown et~al\mbox{.}(2020)]%
        {brown2020language}
\bibfield{author}{\bibinfo{person}{Tom~B Brown}, \bibinfo{person}{Benjamin Mann}, \bibinfo{person}{Nick Ryder}, \bibinfo{person}{Melanie Subbiah}, \bibinfo{person}{Jared Kaplan}, \bibinfo{person}{Prafulla Dhariwal}, \bibinfo{person}{Arvind Neelakantan}, \bibinfo{person}{Pranav Shyam}, \bibinfo{person}{Girish Sastry}, \bibinfo{person}{Amanda Askell}, {et~al\mbox{.}}} \bibinfo{year}{2020}\natexlab{}.
\newblock \showarticletitle{Language models are few-shot learners}. In \bibinfo{booktitle}{\emph{NeurIPS}}.
\newblock


\bibitem[Camburu et~al\mbox{.}(2018)]%
        {camburu2018esnli}
\bibfield{author}{\bibinfo{person}{Oana-Maria Camburu}, \bibinfo{person}{Tim Rockt{\"a}schel}, \bibinfo{person}{Thomas Lukasiewicz}, {and} \bibinfo{person}{Phil Blunsom}.} \bibinfo{year}{2018}\natexlab{}.
\newblock \showarticletitle{e-SNLI: Natural Language Inference with Natural Language Explanations}. In \bibinfo{booktitle}{\emph{NeurIPS}}.
\newblock


\bibitem[Chen et~al\mbox{.}(2023)]%
        {chen2023mcc}
\bibfield{author}{\bibinfo{person}{Hongzhan Chen}, \bibinfo{person}{Siyue Wu}, \bibinfo{person}{Xiaojun Quan}, \bibinfo{person}{Rui Wang}, \bibinfo{person}{Ming Yan}, {and} \bibinfo{person}{Ji Zhang}.} \bibinfo{year}{2023}\natexlab{}.
\newblock \showarticletitle{MCC-KD: Multi-CoT Consistent Knowledge Distillation}. In \bibinfo{booktitle}{\emph{EMNLP Findings}}.
\newblock


\bibitem[Chen et~al\mbox{.}(2024)]%
        {chen2023minprompt}
\bibfield{author}{\bibinfo{person}{Xiusi Chen}, \bibinfo{person}{Jyun-Yu Jiang}, \bibinfo{person}{Wei-Cheng Chang}, \bibinfo{person}{Cho-Jui Hsieh}, \bibinfo{person}{Hsiang-Fu Yu}, {and} \bibinfo{person}{Wei Wang}.} \bibinfo{year}{2024}\natexlab{}.
\newblock \showarticletitle{MinPrompt: Graph-based Minimal Prompt Data Augmentation for Few-shot Question Answering}. In \bibinfo{booktitle}{\emph{ACL}}.
\newblock


\bibitem[Cho and Hariharan(2019)]%
        {cho2019efficacy}
\bibfield{author}{\bibinfo{person}{Jang~Hyun Cho} {and} \bibinfo{person}{Bharath Hariharan}.} \bibinfo{year}{2019}\natexlab{}.
\newblock \showarticletitle{On the efficacy of knowledge distillation}. In \bibinfo{booktitle}{\emph{ICCV}}.
\newblock


\bibitem[Chowdhery et~al\mbox{.}(2023)]%
        {chowdhery2023palm}
\bibfield{author}{\bibinfo{person}{Aakanksha Chowdhery}, \bibinfo{person}{Sharan Narang}, \bibinfo{person}{Jacob Devlin}, \bibinfo{person}{Maarten Bosma}, \bibinfo{person}{Gaurav Mishra}, \bibinfo{person}{Adam Roberts}, \bibinfo{person}{Paul Barham}, \bibinfo{person}{Hyung~Won Chung}, \bibinfo{person}{Charles Sutton}, \bibinfo{person}{Sebastian Gehrmann}, {et~al\mbox{.}}} \bibinfo{year}{2023}\natexlab{}.
\newblock \showarticletitle{PALM: Scaling Language Modeling with Pathways}.
\newblock \bibinfo{journal}{\emph{Journal of Machine Learning Research}} (\bibinfo{year}{2023}).
\newblock


\bibitem[Chung et~al\mbox{.}(2022)]%
        {chung2022scaling}
\bibfield{author}{\bibinfo{person}{Hyung~Won Chung}, \bibinfo{person}{Le Hou}, \bibinfo{person}{Shayne Longpre}, \bibinfo{person}{Barret Zoph}, \bibinfo{person}{Yi Tay}, \bibinfo{person}{William Fedus}, \bibinfo{person}{Yunxuan Li}, \bibinfo{person}{Xuezhi Wang}, \bibinfo{person}{Mostafa Dehghani}, \bibinfo{person}{Siddhartha Brahma}, {et~al\mbox{.}}} \bibinfo{year}{2022}\natexlab{}.
\newblock \showarticletitle{Scaling Instruction-Finetuned Language Models}.
\newblock \bibinfo{journal}{\emph{arXiv preprint arXiv:2210.11416}} (\bibinfo{year}{2022}).
\newblock


\bibitem[Clark et~al\mbox{.}(2018)]%
        {Clark2018ThinkYH}
\bibfield{author}{\bibinfo{person}{Peter Clark}, \bibinfo{person}{Isaac Cowhey}, \bibinfo{person}{Oren Etzioni}, \bibinfo{person}{Tushar Khot}, \bibinfo{person}{Ashish Sabharwal}, \bibinfo{person}{Carissa Schoenick}, {and} \bibinfo{person}{Oyvind Tafjord}.} \bibinfo{year}{2018}\natexlab{}.
\newblock \showarticletitle{Think you have Solved Question Answering? Try ARC, the AI2 Reasoning Challenge}.
\newblock \bibinfo{journal}{\emph{arXiv preprint arXiv:1803.05457}} (\bibinfo{year}{2018}).
\newblock


\bibitem[Dubey et~al\mbox{.}(2024)]%
        {dubey2024llama}
\bibfield{author}{\bibinfo{person}{Abhimanyu Dubey}, \bibinfo{person}{Abhinav Jauhri}, \bibinfo{person}{Abhinav Pandey}, \bibinfo{person}{Abhishek Kadian}, \bibinfo{person}{Ahmad Al-Dahle}, \bibinfo{person}{Aiesha Letman}, \bibinfo{person}{Akhil Mathur}, \bibinfo{person}{Alan Schelten}, \bibinfo{person}{Amy Yang}, \bibinfo{person}{Angela Fan}, {et~al\mbox{.}}} \bibinfo{year}{2024}\natexlab{}.
\newblock \showarticletitle{The llama 3 herd of models}.
\newblock \bibinfo{journal}{\emph{arXiv preprint arXiv:2407.21783}} (\bibinfo{year}{2024}).
\newblock


\bibitem[Eisenstein et~al\mbox{.}(2022)]%
        {eisenstein2022honeststudents}
\bibfield{author}{\bibinfo{person}{Jacob Eisenstein}, \bibinfo{person}{Daniel Andor}, \bibinfo{person}{Bernd Bohnet}, \bibinfo{person}{Michael Collins}, {and} \bibinfo{person}{David Mimno}.} \bibinfo{year}{2022}\natexlab{}.
\newblock \showarticletitle{Honest Students from Untrusted Teachers: Learning an Interpretable Question-Answering Pipeline from a Pretrained Language Model}.
\newblock \bibinfo{journal}{\emph{arXiv preprint arXiv:2210.02498}} (\bibinfo{year}{2022}).
\newblock


\bibitem[Fu et~al\mbox{.}(2023)]%
        {pmlr-v202-fu23d}
\bibfield{author}{\bibinfo{person}{Yao Fu}, \bibinfo{person}{Hao Peng}, \bibinfo{person}{Litu Ou}, \bibinfo{person}{Ashish Sabharwal}, {and} \bibinfo{person}{Tushar Khot}.} \bibinfo{year}{2023}\natexlab{}.
\newblock \showarticletitle{Specializing Smaller Language Models towards Multi-Step Reasoning}. In \bibinfo{booktitle}{\emph{ICML}}.
\newblock


\bibitem[Gou et~al\mbox{.}(2021)]%
        {gou2021knowledge}
\bibfield{author}{\bibinfo{person}{Jianping Gou}, \bibinfo{person}{Baosheng Yu}, \bibinfo{person}{Stephen~J Maybank}, {and} \bibinfo{person}{Dacheng Tao}.} \bibinfo{year}{2021}\natexlab{}.
\newblock \showarticletitle{Knowledge distillation: A survey}.
\newblock \bibinfo{journal}{\emph{International Journal of Computer Vision}} (\bibinfo{year}{2021}).
\newblock


\bibitem[Gu et~al\mbox{.}(2024)]%
        {gu2024minillm}
\bibfield{author}{\bibinfo{person}{Yuxian Gu}, \bibinfo{person}{Li Dong}, \bibinfo{person}{Furu Wei}, {and} \bibinfo{person}{Minlie Huang}.} \bibinfo{year}{2024}\natexlab{}.
\newblock \showarticletitle{MiniLLM: Knowledge distillation of large language models}. In \bibinfo{booktitle}{\emph{The Twelfth International Conference on Learning Representations}}.
\newblock


\bibitem[Gu et~al\mbox{.}(2021)]%
        {gu2021ppt}
\bibfield{author}{\bibinfo{person}{Yuxian Gu}, \bibinfo{person}{Xu Han}, \bibinfo{person}{Zhiyuan Liu}, {and} \bibinfo{person}{Minlie Huang}.} \bibinfo{year}{2021}\natexlab{}.
\newblock \showarticletitle{Ppt: Pre-trained prompt tuning for few-shot learning}.
\newblock \bibinfo{journal}{\emph{arXiv preprint arXiv:2109.04332}} (\bibinfo{year}{2021}).
\newblock


\bibitem[Guo et~al\mbox{.}(2023)]%
        {guo2023boosting}
\bibfield{author}{\bibinfo{person}{Zhichun Guo}, \bibinfo{person}{Chunhui Zhang}, \bibinfo{person}{Yujie Fan}, \bibinfo{person}{Yijun Tian}, \bibinfo{person}{Chuxu Zhang}, {and} \bibinfo{person}{Nitesh~V Chawla}.} \bibinfo{year}{2023}\natexlab{}.
\newblock \showarticletitle{Boosting graph neural networks via adaptive knowledge distillation}. In \bibinfo{booktitle}{\emph{AAAI}}.
\newblock


\bibitem[Hancock et~al\mbox{.}(2019)]%
        {hancock2019feedchatbot}
\bibfield{author}{\bibinfo{person}{Braden Hancock}, \bibinfo{person}{Antoine Bordes}, \bibinfo{person}{Pierre-Emmanuel Mazare}, {and} \bibinfo{person}{Jason Weston}.} \bibinfo{year}{2019}\natexlab{}.
\newblock \showarticletitle{Learning from Dialogue after Deployment: Feed Yourself, Chatbot!}. In \bibinfo{booktitle}{\emph{ACL}}.
\newblock


\bibitem[Hase and Bansal(2022)]%
        {hase2021learningfromexplanations}
\bibfield{author}{\bibinfo{person}{Peter Hase} {and} \bibinfo{person}{Mohit Bansal}.} \bibinfo{year}{2022}\natexlab{}.
\newblock \showarticletitle{When Can Models Learn From Explanations? A Formal Framework for Understanding the Roles of Explanation Data}. In \bibinfo{booktitle}{\emph{ACL Workshop}}.
\newblock


\bibitem[Hinton et~al\mbox{.}(2015)]%
        {hinton_kd}
\bibfield{author}{\bibinfo{person}{Geoffrey Hinton}, \bibinfo{person}{Oriol Vinyals}, \bibinfo{person}{Jeff Dean}, {et~al\mbox{.}}} \bibinfo{year}{2015}\natexlab{}.
\newblock \showarticletitle{Distilling the knowledge in a neural network}.
\newblock \bibinfo{journal}{\emph{arXiv preprint arXiv:1503.02531}} (\bibinfo{year}{2015}).
\newblock


\bibitem[Ho et~al\mbox{.}(2022)]%
        {ho2022largelanguagemodels}
\bibfield{author}{\bibinfo{person}{Namgyu Ho}, \bibinfo{person}{Laura Schmid}, {and} \bibinfo{person}{Se-Young Yun}.} \bibinfo{year}{2022}\natexlab{}.
\newblock \showarticletitle{Large Language Models Are Reasoning Teachers}.
\newblock \bibinfo{journal}{\emph{arXiv preprint arXiv:2212.10071}} (\bibinfo{year}{2022}).
\newblock


\bibitem[Hsieh et~al\mbox{.}(2023)]%
        {hsieh-etal-2023-distilling}
\bibfield{author}{\bibinfo{person}{Cheng-Yu Hsieh}, \bibinfo{person}{Chun-Liang Li}, \bibinfo{person}{Chih-kuan Yeh}, \bibinfo{person}{Hootan Nakhost}, \bibinfo{person}{Yasuhisa Fujii}, \bibinfo{person}{Alex Ratner}, \bibinfo{person}{Ranjay Krishna}, \bibinfo{person}{Chen-Yu Lee}, {and} \bibinfo{person}{Tomas Pfister}.} \bibinfo{year}{2023}\natexlab{}.
\newblock \showarticletitle{Distilling Step-by-Step! Outperforming Larger Language Models with Less Training Data and Smaller Model Sizes}. In \bibinfo{booktitle}{\emph{ACL Findings}}.
\newblock


\bibitem[Hu et~al\mbox{.}(2022)]%
        {hu2022lora}
\bibfield{author}{\bibinfo{person}{Edward~J Hu}, \bibinfo{person}{Yelong Shen}, \bibinfo{person}{Phillip Wallis}, \bibinfo{person}{Zeyuan Allen-Zhu}, \bibinfo{person}{Yuanzhi Li}, \bibinfo{person}{Shean Wang}, \bibinfo{person}{Lu Wang}, {and} \bibinfo{person}{Weizhu Chen}.} \bibinfo{year}{2022}\natexlab{}.
\newblock \showarticletitle{Lo{RA}: Low-Rank Adaptation of Large Language Models}. In \bibinfo{booktitle}{\emph{ICLR}}.
\newblock


\bibitem[Huang and Chang(2022)]%
        {huang2022towards}
\bibfield{author}{\bibinfo{person}{Jie Huang} {and} \bibinfo{person}{Kevin Chen-Chuan Chang}.} \bibinfo{year}{2022}\natexlab{}.
\newblock \showarticletitle{Towards reasoning in large language models: A survey}.
\newblock \bibinfo{journal}{\emph{arXiv preprint arXiv:2212.10403}} (\bibinfo{year}{2022}).
\newblock


\bibitem[Huang et~al\mbox{.}(2023)]%
        {huang2023large}
\bibfield{author}{\bibinfo{person}{Jiaxin Huang}, \bibinfo{person}{Shixiang Gu}, \bibinfo{person}{Le Hou}, \bibinfo{person}{Yuexin Wu}, \bibinfo{person}{Xuezhi Wang}, \bibinfo{person}{Hongkun Yu}, {and} \bibinfo{person}{Jiawei Han}.} \bibinfo{year}{2023}\natexlab{}.
\newblock \showarticletitle{Large Language Models Can Self-Improve}. In \bibinfo{booktitle}{\emph{Proceedings of the 2023 Conference on Empirical Methods in Natural Language Processing}}. \bibinfo{pages}{1051--1068}.
\newblock


\bibitem[Huang et~al\mbox{.}(2022)]%
        {huang2022languagemodelsselfimprove}
\bibfield{author}{\bibinfo{person}{Jiaxin Huang}, \bibinfo{person}{Shixiang~Shane Gu}, \bibinfo{person}{Le Hou}, \bibinfo{person}{Yuexin Wu}, \bibinfo{person}{Xuezhi Wang}, \bibinfo{person}{Hongkun Yu}, {and} \bibinfo{person}{Jiawei Han}.} \bibinfo{year}{2022}\natexlab{}.
\newblock \showarticletitle{Large language models can self-improve}.
\newblock \bibinfo{journal}{\emph{arXiv preprint arXiv:2210.11610}} (\bibinfo{year}{2022}).
\newblock


\bibitem[Ji et~al\mbox{.}(2023)]%
        {ji2023hallucinationnlg}
\bibfield{author}{\bibinfo{person}{Z Ji}, \bibinfo{person}{N Lee}, \bibinfo{person}{R Frieske}, \bibinfo{person}{T Yu}, \bibinfo{person}{D Su}, \bibinfo{person}{Y Xu}, \bibinfo{person}{E Ishii}, \bibinfo{person}{Y~J Bang}, \bibinfo{person}{A Madotto}, {and} \bibinfo{person}{P Fung}.} \bibinfo{year}{2023}\natexlab{}.
\newblock \showarticletitle{Survey of Hallucination in Natural Language Generation}.
\newblock \bibinfo{journal}{\emph{Comput. Surveys}} (\bibinfo{year}{2023}).
\newblock


\bibitem[Jin et~al\mbox{.}(2019)]%
        {jin2019pubmedqa}
\bibfield{author}{\bibinfo{person}{Qiao Jin}, \bibinfo{person}{Bhuwan Dhingra}, \bibinfo{person}{Zhengping Liu}, \bibinfo{person}{William Cohen}, {and} \bibinfo{person}{Xinghua Lu}.} \bibinfo{year}{2019}\natexlab{}.
\newblock \showarticletitle{PubMedQA: A Dataset for Biomedical Research Question Answering}. In \bibinfo{booktitle}{\emph{EMNLP}}.
\newblock


\bibitem[Kaplan et~al\mbox{.}(2020)]%
        {kaplan2020scalinglaws}
\bibfield{author}{\bibinfo{person}{J Kaplan}, \bibinfo{person}{S McCandlish}, \bibinfo{person}{T Henighan}, {et~al\mbox{.}}} \bibinfo{year}{2020}\natexlab{}.
\newblock \showarticletitle{Scaling laws for neural language models}.
\newblock \bibinfo{journal}{\emph{arXiv preprint arXiv:2001.08361}} (\bibinfo{year}{2020}).
\newblock


\bibitem[Ko et~al\mbox{.}({[n.\,d.]})]%
        {kodistillm}
\bibfield{author}{\bibinfo{person}{Jongwoo Ko}, \bibinfo{person}{Sungnyun Kim}, \bibinfo{person}{Tianyi Chen}, {and} \bibinfo{person}{Se-Young Yun}.} \bibinfo{year}{[n.\,d.]}\natexlab{}.
\newblock \showarticletitle{DistiLLM: Towards Streamlined Distillation for Large Language Models}. In \bibinfo{booktitle}{\emph{Forty-first International Conference on Machine Learning}}.
\newblock


\bibitem[Kojima et~al\mbox{.}(2022)]%
        {kojima2022zero}
\bibfield{author}{\bibinfo{person}{Takeshi Kojima}, \bibinfo{person}{Shixiang~(Shane) Gu}, \bibinfo{person}{Machel Reid}, \bibinfo{person}{Yutaka Matsuo}, {and} \bibinfo{person}{Yusuke Iwasawa}.} \bibinfo{year}{2022}\natexlab{}.
\newblock \showarticletitle{Large Language Models are Zero-Shot Reasoners}. In \bibinfo{booktitle}{\emph{NeurIPS}}.
\newblock


\bibitem[Lester et~al\mbox{.}(2021)]%
        {lester2021powerofscale}
\bibfield{author}{\bibinfo{person}{B Lester}, \bibinfo{person}{R Al-Rfou}, {and} \bibinfo{person}{N Constant}.} \bibinfo{year}{2021}\natexlab{}.
\newblock \showarticletitle{The Power of Scale for Parameter-Efficient Prompt Tuning}. In \bibinfo{booktitle}{\emph{EMNLP}}.
\newblock


\bibitem[Li et~al\mbox{.}(2023)]%
        {li2023symbolicchain}
\bibfield{author}{\bibinfo{person}{Liunian~Harold Li} {et~al\mbox{.}}} \bibinfo{year}{2023}\natexlab{}.
\newblock \showarticletitle{Symbolic Chain-of-Thought Distillation: Small Models Can Also "Think" Step-by-Step}.
\newblock \bibinfo{journal}{\emph{arXiv preprint arXiv:2306.14050}} (\bibinfo{year}{2023}).
\newblock


\bibitem[Li and Liang(2021)]%
        {li2021prefixtuning}
\bibfield{author}{\bibinfo{person}{X~L Li} {and} \bibinfo{person}{P Liang}.} \bibinfo{year}{2021}\natexlab{}.
\newblock \showarticletitle{Prefix-Tuning: Optimizing Continuous Prompts for Generation}. In \bibinfo{booktitle}{\emph{ACL}}.
\newblock


\bibitem[Lin et~al\mbox{.}(2021)]%
        {lin-etal-2021-riddlesense}
\bibfield{author}{\bibinfo{person}{Bill~Yuchen Lin}, \bibinfo{person}{Ziyi Wu}, \bibinfo{person}{Yichi Yang}, \bibinfo{person}{Dong-Ho Lee}, {and} \bibinfo{person}{Xiang Ren}.} \bibinfo{year}{2021}\natexlab{}.
\newblock \showarticletitle{RiddleSense: Reasoning about Riddle Questions Featuring Linguistic Creativity and Commonsense Knowledge}. In \bibinfo{booktitle}{\emph{ACL Findings}}.
\newblock


\bibitem[Liu et~al\mbox{.}(2023)]%
        {liu2023mind}
\bibfield{author}{\bibinfo{person}{Weize Liu}, \bibinfo{person}{Guocong Li}, \bibinfo{person}{Kai Zhang}, \bibinfo{person}{Bang Du}, \bibinfo{person}{Qiyuan Chen}, \bibinfo{person}{Xuming Hu}, \bibinfo{person}{Hongxia Xu}, \bibinfo{person}{Jintai Chen}, {and} \bibinfo{person}{Jian Wu}.} \bibinfo{year}{2023}\natexlab{}.
\newblock \showarticletitle{Mind's Mirror: Distilling Self-Evaluation Capability and Comprehensive Thinking from Large Language Models}.
\newblock \bibinfo{journal}{\emph{arXiv preprint arXiv:2311.09214}} (\bibinfo{year}{2023}).
\newblock


\bibitem[Liu et~al\mbox{.}(2021)]%
        {liu2021p}
\bibfield{author}{\bibinfo{person}{Xiao Liu}, \bibinfo{person}{Kaixuan Ji}, \bibinfo{person}{Yicheng Fu}, \bibinfo{person}{Weng~Lam Tam}, \bibinfo{person}{Zhengxiao Du}, \bibinfo{person}{Zhilin Yang}, {and} \bibinfo{person}{Jie Tang}.} \bibinfo{year}{2021}\natexlab{}.
\newblock \showarticletitle{P-tuning v2: Prompt tuning can be comparable to fine-tuning universally across scales and tasks}.
\newblock \bibinfo{journal}{\emph{arXiv preprint arXiv:2110.07602}} (\bibinfo{year}{2021}).
\newblock


\bibitem[Liu et~al\mbox{.}(2024a)]%
        {sku}
\bibfield{author}{\bibinfo{person}{Zheyuan Liu}, \bibinfo{person}{Guangyao Dou}, \bibinfo{person}{Zhaoxuan Tan}, \bibinfo{person}{Yijun Tian}, {and} \bibinfo{person}{Meng Jiang}.} \bibinfo{year}{2024}\natexlab{a}.
\newblock \showarticletitle{Towards Safer Large Language Models through Machine Unlearning}.
\newblock \bibinfo{journal}{\emph{arXiv preprint arXiv:2402.10058}} (\bibinfo{year}{2024}).
\newblock


\bibitem[Liu et~al\mbox{.}(2024b)]%
        {prompt_llm_for_graph}
\bibfield{author}{\bibinfo{person}{Zheyuan Liu}, \bibinfo{person}{Xiaoxin He}, \bibinfo{person}{Yijun Tian}, {and} \bibinfo{person}{Nitesh~V Chawla}.} \bibinfo{year}{2024}\natexlab{b}.
\newblock \showarticletitle{Can we soft prompt LLMs for graph learning tasks?}. In \bibinfo{booktitle}{\emph{WWW}}.
\newblock


\bibitem[Lu et~al\mbox{.}(2022)]%
        {lu2022learntoexplain}
\bibfield{author}{\bibinfo{person}{P Lu}, \bibinfo{person}{S Mishra}, \bibinfo{person}{T Xia}, \bibinfo{person}{L Qiu}, \bibinfo{person}{K-W Chang}, \bibinfo{person}{S-C Zhu}, \bibinfo{person}{O Tafjord}, \bibinfo{person}{P Clark}, {and} \bibinfo{person}{A Kalyan}.} \bibinfo{year}{2022}\natexlab{}.
\newblock \showarticletitle{Learn to Explain: Multimodal Reasoning via Thought Chains for Science Question Answering}. In \bibinfo{booktitle}{\emph{NeurIPS}}.
\newblock


\bibitem[Magister et~al\mbox{.}(2022)]%
        {magister2022teachingsmall}
\bibfield{author}{\bibinfo{person}{Lucie~Charlotte Magister}, \bibinfo{person}{Jonathan Mallinson}, \bibinfo{person}{Jakub Adamek}, \bibinfo{person}{Eric Malmi}, {and} \bibinfo{person}{Aliaksei Severyn}.} \bibinfo{year}{2022}\natexlab{}.
\newblock \showarticletitle{Teaching Small Language Models to Reason}.
\newblock \bibinfo{journal}{\emph{arXiv preprint arXiv:2212.08410}} (\bibinfo{year}{2022}).
\newblock


\bibitem[Mihaylov et~al\mbox{.}(2018)]%
        {OpenBookQA2018}
\bibfield{author}{\bibinfo{person}{Todor Mihaylov}, \bibinfo{person}{Peter Clark}, \bibinfo{person}{Tushar Khot}, {and} \bibinfo{person}{Ashish Sabharwal}.} \bibinfo{year}{2018}\natexlab{}.
\newblock \showarticletitle{Can a Suit of Armor Conduct Electricity? A New Dataset for Open Book Question Answering}. In \bibinfo{booktitle}{\emph{EMNLP}}.
\newblock


\bibitem[Narang et~al\mbox{.}(2020)]%
        {narang2020wt5}
\bibfield{author}{\bibinfo{person}{Sharan Narang}, \bibinfo{person}{Colin Raffel}, \bibinfo{person}{Katherine Lee}, \bibinfo{person}{Adam Roberts}, \bibinfo{person}{Noah Fiedel}, {and} \bibinfo{person}{Karishma Malkan}.} \bibinfo{year}{2020}\natexlab{}.
\newblock \showarticletitle{WT5?! Training Text-to-Text Models to Explain Their Predictions}.
\newblock \bibinfo{journal}{\emph{arXiv preprint arXiv:2004.14546}} (\bibinfo{year}{2020}).
\newblock


\bibitem[Pruthi et~al\mbox{.}(2022)]%
        {pruthi2022evaluatingexplanations}
\bibfield{author}{\bibinfo{person}{Danish Pruthi}, \bibinfo{person}{Rachit Bansal}, \bibinfo{person}{Bhuwan Dhingra}, \bibinfo{person}{Livio~Baldini Soares}, \bibinfo{person}{Michael Collins}, \bibinfo{person}{Zachary~C Lipton}, \bibinfo{person}{Graham Neubig}, {and} \bibinfo{person}{William~W Cohen}.} \bibinfo{year}{2022}\natexlab{}.
\newblock \showarticletitle{Evaluating Explanations: How Much Do Explanations from the Teacher Aid Students?}. In \bibinfo{booktitle}{\emph{ACL}}.
\newblock


\bibitem[Raffel et~al\mbox{.}(2020)]%
        {raffel2020exploringlimits}
\bibfield{author}{\bibinfo{person}{C Raffel}, \bibinfo{person}{N Shazeer}, \bibinfo{person}{A Roberts}, {et~al\mbox{.}}} \bibinfo{year}{2020}\natexlab{}.
\newblock \showarticletitle{Exploring the Limits of Transfer Learning with a Unified Text-to-Text Transformer}.
\newblock \bibinfo{journal}{\emph{Journal of Machine Learning Research}} (\bibinfo{year}{2020}).
\newblock


\bibitem[Rajani et~al\mbox{.}(2019)]%
        {rajani2019explainyourself}
\bibfield{author}{\bibinfo{person}{Nazneen~Fatema Rajani}, \bibinfo{person}{Bryan McCann}, \bibinfo{person}{Caiming Xiong}, {and} \bibinfo{person}{Richard Socher}.} \bibinfo{year}{2019}\natexlab{}.
\newblock \showarticletitle{Explain Yourself! Leveraging Language Models for Commonsense Reasoning}. In \bibinfo{booktitle}{\emph{ACL}}.
\newblock


\bibitem[Rao et~al\mbox{.}(2023)]%
        {rao2023makes}
\bibfield{author}{\bibinfo{person}{Kavel Rao}, \bibinfo{person}{Liwei Jiang}, \bibinfo{person}{Valentina Pyatkin}, \bibinfo{person}{Yuling Gu}, \bibinfo{person}{Niket Tandon}, \bibinfo{person}{Nouha Dziri}, \bibinfo{person}{Faeze Brahman}, {and} \bibinfo{person}{Yejin Choi}.} \bibinfo{year}{2023}\natexlab{}.
\newblock \showarticletitle{What Makes it Ok to Set a Fire? Iterative Self-distillation of Contexts and Rationales for Disambiguating Defeasible Social and Moral Situations}. In \bibinfo{booktitle}{\emph{EMNLP Findings}}.
\newblock


\bibitem[Ross et~al\mbox{.}(2017)]%
        {ross2017rightreasons}
\bibfield{author}{\bibinfo{person}{Andrew~Slavin Ross}, \bibinfo{person}{Michael~C Hughes}, {and} \bibinfo{person}{Finale Doshi-Velez}.} \bibinfo{year}{2017}\natexlab{}.
\newblock \showarticletitle{Right for the Right Reasons: Training Differentiable Models by Constraining Their Explanations}.
\newblock \bibinfo{journal}{\emph{arXiv preprint arXiv:1703.03717}} (\bibinfo{year}{2017}).
\newblock


\bibitem[Sanh et~al\mbox{.}(2019)]%
        {sanh2019distilbert}
\bibfield{author}{\bibinfo{person}{Victor Sanh}, \bibinfo{person}{Lysandre Debut}, \bibinfo{person}{Julien Chaumond}, {and} \bibinfo{person}{Thomas Wolf}.} \bibinfo{year}{2019}\natexlab{}.
\newblock \showarticletitle{DistilBERT, a distilled version of BERT: smaller, faster, cheaper and lighter}.
\newblock \bibinfo{journal}{\emph{arXiv preprint arXiv:1910.01108}} (\bibinfo{year}{2019}).
\newblock


\bibitem[Shi et~al\mbox{.}(2023)]%
        {shi2023mededit}
\bibfield{author}{\bibinfo{person}{Yucheng Shi}, \bibinfo{person}{Shaochen Xu}, {et~al\mbox{.}}} \bibinfo{year}{2023}\natexlab{}.
\newblock \showarticletitle{Mededit: Model Editing for Medical Question Answering with External Knowledge Bases}.
\newblock \bibinfo{journal}{\emph{arXiv preprint arXiv:2309.16035}} (\bibinfo{year}{2023}).
\newblock


\bibitem[Smith et~al\mbox{.}(2022)]%
        {smith2022deepspeedmegatron}
\bibfield{author}{\bibinfo{person}{Shaden Smith}, \bibinfo{person}{Mostofa Patwary}, \bibinfo{person}{Brandon Norick}, \bibinfo{person}{Patrick LeGresley}, \bibinfo{person}{Samyam Rajbhandari}, \bibinfo{person}{Jared Casper}, \bibinfo{person}{Zhun Liu}, \bibinfo{person}{Shrimai Prabhumoye}, \bibinfo{person}{George Zerveas}, \bibinfo{person}{Vijay Korthikanti}, \bibinfo{person}{Elton Zhang}, \bibinfo{person}{Rewon Child}, \bibinfo{person}{Reza~Yazdani Aminabadi}, \bibinfo{person}{Julie Bernauer}, \bibinfo{person}{Xia Song}, \bibinfo{person}{Mohammad Shoeybi}, \bibinfo{person}{Yuxiong He}, \bibinfo{person}{Michael Houston}, \bibinfo{person}{Saurabh Tiwary}, {and} \bibinfo{person}{Bryan Catanzaro}.} \bibinfo{year}{2022}\natexlab{}.
\newblock \showarticletitle{Using DeepSpeed and Megatron to Train Megatron-Turing NLG 530B, a Large-Scale Generative Language Model}.
\newblock \bibinfo{journal}{\emph{arXiv preprint arXiv:2201.11990}} (\bibinfo{year}{2022}).
\newblock


\bibitem[Tan et~al\mbox{.}(2024)]%
        {oppu}
\bibfield{author}{\bibinfo{person}{Zhaoxuan Tan}, \bibinfo{person}{Qingkai Zeng}, \bibinfo{person}{Yijun Tian}, \bibinfo{person}{Zheyuan Liu}, \bibinfo{person}{Bing Yin}, {and} \bibinfo{person}{Meng Jiang}.} \bibinfo{year}{2024}\natexlab{}.
\newblock \showarticletitle{Democratizing Large Language Models via Personalized Parameter-Efficient Fine-tuning}.
\newblock \bibinfo{journal}{\emph{arXiv preprint arXiv:2402.04401}} (\bibinfo{year}{2024}).
\newblock


\bibitem[Taori et~al\mbox{.}(2023)]%
        {alpaca}
\bibfield{author}{\bibinfo{person}{Rohan Taori}, \bibinfo{person}{Ishaan Gulrajani}, \bibinfo{person}{Tianyi Zhang}, \bibinfo{person}{Yann Dubois}, \bibinfo{person}{Xuechen Li}, \bibinfo{person}{Carlos Guestrin}, \bibinfo{person}{Percy Liang}, {and} \bibinfo{person}{Tatsunori~B Hashimoto}.} \bibinfo{year}{2023}\natexlab{}.
\newblock \showarticletitle{Stanford Alpaca: An Instruction-following LLaMA model}.
\newblock \bibinfo{journal}{\emph{GitHub repository}} (\bibinfo{year}{2023}).
\newblock


\bibitem[Team et~al\mbox{.}(2024)]%
        {team2024gemma}
\bibfield{author}{\bibinfo{person}{Gemma Team}, \bibinfo{person}{Morgane Riviere}, \bibinfo{person}{Shreya Pathak}, \bibinfo{person}{Pier~Giuseppe Sessa}, \bibinfo{person}{Cassidy Hardin}, \bibinfo{person}{Surya Bhupatiraju}, \bibinfo{person}{L{\'e}onard Hussenot}, \bibinfo{person}{Thomas Mesnard}, \bibinfo{person}{Bobak Shahriari}, \bibinfo{person}{Alexandre Ram{\'e}}, {et~al\mbox{.}}} \bibinfo{year}{2024}\natexlab{}.
\newblock \showarticletitle{Gemma 2: Improving open language models at a practical size}.
\newblock \bibinfo{journal}{\emph{arXiv preprint arXiv:2408.00118}} (\bibinfo{year}{2024}).
\newblock


\bibitem[Tian et~al\mbox{.}(2023a)]%
        {kd_on_graph_survey}
\bibfield{author}{\bibinfo{person}{Yijun Tian}, \bibinfo{person}{Shichao Pei}, \bibinfo{person}{Xiangliang Zhang}, \bibinfo{person}{Chuxu Zhang}, {and} \bibinfo{person}{Nitesh~V Chawla}.} \bibinfo{year}{2023}\natexlab{a}.
\newblock \showarticletitle{Knowledge Distillation on Graphs: A Survey}.
\newblock \bibinfo{journal}{\emph{arXiv preprint arXiv:2302.00219}} (\bibinfo{year}{2023}).
\newblock


\bibitem[Tian et~al\mbox{.}(2024)]%
        {gnp}
\bibfield{author}{\bibinfo{person}{Yijun Tian}, \bibinfo{person}{Huan Song}, \bibinfo{person}{Zichen Wang}, \bibinfo{person}{Haozhu Wang}, \bibinfo{person}{Ziqing Hu}, \bibinfo{person}{Fang Wang}, \bibinfo{person}{Nitesh~V Chawla}, {and} \bibinfo{person}{Panpan Xu}.} \bibinfo{year}{2024}\natexlab{}.
\newblock \showarticletitle{Graph neural prompting with large language models}. In \bibinfo{booktitle}{\emph{AAAI}}.
\newblock


\bibitem[Tian et~al\mbox{.}(2023b)]%
        {nosmog}
\bibfield{author}{\bibinfo{person}{Yijun Tian}, \bibinfo{person}{Chuxu Zhang}, \bibinfo{person}{Zhichun Guo}, \bibinfo{person}{Xiangliang Zhang}, {and} \bibinfo{person}{Nitesh~V Chawla}.} \bibinfo{year}{2023}\natexlab{b}.
\newblock \showarticletitle{Learning MLPs on Graphs: A Unified View of Effectiveness, Robustness, and Efficiency}. In \bibinfo{booktitle}{\emph{ICLR}}.
\newblock


\bibitem[Touvron et~al\mbox{.}(2023a)]%
        {touvron2023llama}
\bibfield{author}{\bibinfo{person}{Hugo Touvron}, \bibinfo{person}{Thibaut Lavril}, {et~al\mbox{.}}} \bibinfo{year}{2023}\natexlab{a}.
\newblock \showarticletitle{Llama: Open and Efficient Foundation Language Models}.
\newblock \bibinfo{journal}{\emph{arXiv preprint arXiv:2302.13971}} (\bibinfo{year}{2023}).
\newblock


\bibitem[Touvron et~al\mbox{.}(2023b)]%
        {touvron2023llama2}
\bibfield{author}{\bibinfo{person}{Hugo Touvron}, \bibinfo{person}{Louis Martin}, {et~al\mbox{.}}} \bibinfo{year}{2023}\natexlab{b}.
\newblock \showarticletitle{Llama 2: Open Foundation and Fine-Tuned Chat Models}.
\newblock \bibinfo{journal}{\emph{arXiv preprint arXiv:2307.09288}} (\bibinfo{year}{2023}).
\newblock


\bibitem[Tsatsaronis et~al\mbox{.}(2015)]%
        {Tsatsaronis2015BIOASQ}
\bibfield{author}{\bibinfo{person}{George Tsatsaronis}, \bibinfo{person}{Georgios Balikas}, \bibinfo{person}{Prodromos Malakasiotis}, {et~al\mbox{.}}} \bibinfo{year}{2015}\natexlab{}.
\newblock \showarticletitle{An overview of the BIOASQ large-scale biomedical semantic indexing and question answering competition}.
\newblock \bibinfo{journal}{\emph{BMC Bioinformatics}} (\bibinfo{year}{2015}).
\newblock


\bibitem[Turpin et~al\mbox{.}(2024)]%
        {turpin2024language}
\bibfield{author}{\bibinfo{person}{Miles Turpin}, \bibinfo{person}{Julian Michael}, \bibinfo{person}{Ethan Perez}, {and} \bibinfo{person}{Samuel Bowman}.} \bibinfo{year}{2024}\natexlab{}.
\newblock \showarticletitle{Language models don't always say what they think: unfaithful explanations in chain-of-thought prompting}.
\newblock \bibinfo{journal}{\emph{Advances in Neural Information Processing Systems}}  \bibinfo{volume}{36} (\bibinfo{year}{2024}).
\newblock


\bibitem[Wan et~al\mbox{.}(2023)]%
        {wan2023efficient}
\bibfield{author}{\bibinfo{person}{Zhongwei Wan}, \bibinfo{person}{Xin Wang}, \bibinfo{person}{Che Liu}, \bibinfo{person}{Samiul Alam}, \bibinfo{person}{Yu Zheng}, \bibinfo{person}{Zhongnan Qu}, \bibinfo{person}{Shen Yan}, \bibinfo{person}{Yi Zhu}, \bibinfo{person}{Quanlu Zhang}, \bibinfo{person}{Mosharaf Chowdhury}, {et~al\mbox{.}}} \bibinfo{year}{2023}\natexlab{}.
\newblock \showarticletitle{Efficient large language models: A survey}.
\newblock \bibinfo{journal}{\emph{arXiv preprint arXiv:2312.03863}}  \bibinfo{volume}{1} (\bibinfo{year}{2023}).
\newblock


\bibitem[Wang et~al\mbox{.}(2022b)]%
        {wang2022no}
\bibfield{author}{\bibinfo{person}{Chaozheng Wang}, \bibinfo{person}{Yuanhang Yang}, \bibinfo{person}{Cuiyun Gao}, \bibinfo{person}{Yun Peng}, \bibinfo{person}{Hongyu Zhang}, {and} \bibinfo{person}{Michael~R Lyu}.} \bibinfo{year}{2022}\natexlab{b}.
\newblock \showarticletitle{No more fine-tuning? an experimental evaluation of prompt tuning in code intelligence}. In \bibinfo{booktitle}{\emph{Proceedings of the 30th ACM joint European software engineering conference and symposium on the foundations of software engineering}}. \bibinfo{pages}{382--394}.
\newblock


\bibitem[Wang et~al\mbox{.}(2024)]%
        {wang2024roselora}
\bibfield{author}{\bibinfo{person}{Haoyu Wang}, \bibinfo{person}{Tianci Liu}, \bibinfo{person}{Ruirui Li}, \bibinfo{person}{Monica Cheng}, \bibinfo{person}{Tuo Zhao}, {and} \bibinfo{person}{Jing Gao}.} \bibinfo{year}{2024}\natexlab{}.
\newblock \showarticletitle{Roselora: Row and column-wise sparse low-rank adaptation of pre-trained language model for knowledge editing and fine-tuning}.
\newblock \bibinfo{journal}{\emph{arXiv preprint arXiv:2406.10777}} (\bibinfo{year}{2024}).
\newblock


\bibitem[Wang et~al\mbox{.}(2023)]%
        {wang2023hadskip}
\bibfield{author}{\bibinfo{person}{Haoyu Wang}, \bibinfo{person}{Yaqing Wang}, \bibinfo{person}{Tianci Liu}, \bibinfo{person}{Tuo Zhao}, {and} \bibinfo{person}{Jing Gao}.} \bibinfo{year}{2023}\natexlab{}.
\newblock \showarticletitle{HadSkip: Homotopic and Adaptive Layer Skipping of Pre-trained Language Models for Efficient Inference}. In \bibinfo{booktitle}{\emph{EMNLP Findings}}.
\newblock


\bibitem[Wang et~al\mbox{.}(2022a)]%
        {wang2022pinto}
\bibfield{author}{\bibinfo{person}{Peifeng Wang}, \bibinfo{person}{Aaron Chan}, \bibinfo{person}{Filip Ilievski}, \bibinfo{person}{Muhao Chen}, {and} \bibinfo{person}{Xiang Ren}.} \bibinfo{year}{2022}\natexlab{a}.
\newblock \showarticletitle{PINTO: Faithful Language Reasoning Using Prompt-Generated Rationales}. In \bibinfo{booktitle}{\emph{ICLR}}.
\newblock


\bibitem[Wang et~al\mbox{.}(2021)]%
        {wang2021reduce}
\bibfield{author}{\bibinfo{person}{Shuohang Wang}, \bibinfo{person}{Yang Liu}, \bibinfo{person}{Yichong Xu}, \bibinfo{person}{Chenguang Zhu}, {and} \bibinfo{person}{Michael Zeng}.} \bibinfo{year}{2021}\natexlab{}.
\newblock \showarticletitle{Want To Reduce Labeling Cost? {GPT}-3 Can Help}. In \bibinfo{booktitle}{\emph{EMNLP Findings}}.
\newblock


\bibitem[Wei et~al\mbox{.}(2022a)]%
        {wei2022emergentabilities}
\bibfield{author}{\bibinfo{person}{J Wei}, \bibinfo{person}{Y Tay}, \bibinfo{person}{R Bommasani}, {et~al\mbox{.}}} \bibinfo{year}{2022}\natexlab{a}.
\newblock \showarticletitle{Emergent Abilities of Large Language Models}.
\newblock \bibinfo{journal}{\emph{arXiv preprint arXiv:2206.07682}} (\bibinfo{year}{2022}).
\newblock


\bibitem[Wei et~al\mbox{.}(2022b)]%
        {wei2022chainofthought}
\bibfield{author}{\bibinfo{person}{Jason Wei}, \bibinfo{person}{Xuezhi Wang}, {et~al\mbox{.}}} \bibinfo{year}{2022}\natexlab{b}.
\newblock \showarticletitle{Chain-of-Thought Prompting Elicits Reasoning in Large Language Models}. In \bibinfo{booktitle}{\emph{NeurIPS}}.
\newblock


\bibitem[Wei et~al\mbox{.}(2024)]%
        {wei2024llmrec}
\bibfield{author}{\bibinfo{person}{W Wei}, \bibinfo{person}{X Ren}, \bibinfo{person}{J Tang}, \bibinfo{person}{Q Wang}, \bibinfo{person}{L Su}, \bibinfo{person}{S Cheng}, \bibinfo{person}{J Wang}, \bibinfo{person}{D Yin}, {and} \bibinfo{person}{C Huang}.} \bibinfo{year}{2024}\natexlab{}.
\newblock \showarticletitle{LLMRec: Large Language Models with Graph Augmentation for Recommendation}. In \bibinfo{booktitle}{\emph{WSDM}}.
\newblock


\bibitem[Wiegreffe et~al\mbox{.}(2021)]%
        {wiegreffe2021measuringassociation}
\bibfield{author}{\bibinfo{person}{Sarah Wiegreffe}, \bibinfo{person}{Ana Marasovic}, {and} \bibinfo{person}{Noah~A Smith}.} \bibinfo{year}{2021}\natexlab{}.
\newblock \showarticletitle{Measuring Association Between Labels and Free-Text Rationales}. In \bibinfo{booktitle}{\emph{EMNLP}}.
\newblock


\bibitem[Workshop et~al\mbox{.}(2022)]%
        {scao2022bloom}
\bibfield{author}{\bibinfo{person}{BigScience Workshop}, \bibinfo{person}{Teven~Le Scao}, \bibinfo{person}{Angela Fan}, \bibinfo{person}{Christopher Akiki}, \bibinfo{person}{Ellie Pavlick}, \bibinfo{person}{Suzana Ili{\'c}}, \bibinfo{person}{Daniel Hesslow}, \bibinfo{person}{Roman Castagn{\'e}}, \bibinfo{person}{Alexandra~Sasha Luccioni}, \bibinfo{person}{Fran{\c{c}}ois Yvon}, {et~al\mbox{.}}} \bibinfo{year}{2022}\natexlab{}.
\newblock \showarticletitle{Bloom: A 176b-parameter open-access multilingual language model}.
\newblock \bibinfo{journal}{\emph{arXiv preprint arXiv:2211.05100}} (\bibinfo{year}{2022}).
\newblock


\bibitem[Wu et~al\mbox{.}(2023)]%
        {wu2023survey}
\bibfield{author}{\bibinfo{person}{Likang Wu}, \bibinfo{person}{Zhi Zheng}, \bibinfo{person}{Zhaopeng Qiu}, \bibinfo{person}{Hao Wang}, \bibinfo{person}{Hongchao Gu}, \bibinfo{person}{Tingjia Shen}, \bibinfo{person}{Chuan Qin}, \bibinfo{person}{Chen Zhu}, \bibinfo{person}{Hengshu Zhu}, \bibinfo{person}{Qi Liu}, {et~al\mbox{.}}} \bibinfo{year}{2023}\natexlab{}.
\newblock \showarticletitle{A Survey on Large Language Models for Recommendation}.
\newblock \bibinfo{journal}{\emph{arXiv preprint arXiv:2305.19860}} (\bibinfo{year}{2023}).
\newblock


\bibitem[Xu et~al\mbox{.}(2024)]%
        {xu2024survey}
\bibfield{author}{\bibinfo{person}{Xiaohan Xu}, \bibinfo{person}{Ming Li}, \bibinfo{person}{Chongyang Tao}, \bibinfo{person}{Tao Shen}, \bibinfo{person}{Reynold Cheng}, \bibinfo{person}{Jinyang Li}, \bibinfo{person}{Can Xu}, \bibinfo{person}{Dacheng Tao}, {and} \bibinfo{person}{Tianyi Zhou}.} \bibinfo{year}{2024}\natexlab{}.
\newblock \showarticletitle{A survey on knowledge distillation of large language models}.
\newblock \bibinfo{journal}{\emph{arXiv preprint arXiv:2402.13116}} (\bibinfo{year}{2024}).
\newblock


\bibitem[Zaidan et~al\mbox{.}(2007)]%
        {zaidan2007annotatorrationales}
\bibfield{author}{\bibinfo{person}{Omar Zaidan}, \bibinfo{person}{Jason Eisner}, {and} \bibinfo{person}{Christine Piatko}.} \bibinfo{year}{2007}\natexlab{}.
\newblock \showarticletitle{Using “Annotator Rationales” to Improve Machine Learning for Text Categorization}. In \bibinfo{booktitle}{\emph{NAACL}}.
\newblock


\bibitem[Zelikman et~al\mbox{.}(2023)]%
        {zelikman2023generating}
\bibfield{author}{\bibinfo{person}{Eric Zelikman}, \bibinfo{person}{Wanjing Ma}, \bibinfo{person}{Jasmine Tran}, \bibinfo{person}{Diyi Yang}, \bibinfo{person}{Jason Yeatman}, {and} \bibinfo{person}{Nick Haber}.} \bibinfo{year}{2023}\natexlab{}.
\newblock \showarticletitle{Generating and Evaluating Tests for K-12 Students with Language Model Simulations: A Case Study on Sentence Reading Efficiency}. In \bibinfo{booktitle}{\emph{EMNLP}}.
\newblock


\bibitem[Zelikman et~al\mbox{.}(2022)]%
        {zelikman2022star}
\bibfield{author}{\bibinfo{person}{Eric Zelikman}, \bibinfo{person}{Yuhuai Wu}, \bibinfo{person}{Jesse Mu}, {and} \bibinfo{person}{Noah Goodman}.} \bibinfo{year}{2022}\natexlab{}.
\newblock \showarticletitle{Star: Bootstrapping Reasoning with Reasoning}. In \bibinfo{booktitle}{\emph{NeurIPS}}.
\newblock


\bibitem[Zhang et~al\mbox{.}(2016)]%
        {zhang2016rationalecnn}
\bibfield{author}{\bibinfo{person}{Ye Zhang}, \bibinfo{person}{Iain Marshall}, {and} \bibinfo{person}{Byron~C Wallace}.} \bibinfo{year}{2016}\natexlab{}.
\newblock \showarticletitle{Rationale-Augmented Convolutional Neural Networks for Text Classification}. In \bibinfo{booktitle}{\emph{EMNLP}}.
\newblock


\bibitem[Zhang et~al\mbox{.}({[n.\,d.]})]%
        {zhangmultimodal}
\bibfield{author}{\bibinfo{person}{Zhuosheng Zhang}, \bibinfo{person}{Aston Zhang}, \bibinfo{person}{Mu Li}, \bibinfo{person}{George Karypis}, \bibinfo{person}{Alex Smola}, {et~al\mbox{.}}} \bibinfo{year}{[n.\,d.]}\natexlab{}.
\newblock \showarticletitle{Multimodal Chain-of-Thought Reasoning in Language Models}.
\newblock \bibinfo{journal}{\emph{Transactions on Machine Learning Research}} (\bibinfo{year}{[n.\,d.]}).
\newblock


\bibitem[Zhao et~al\mbox{.}(2023)]%
        {zhao2023surveyllms}
\bibfield{author}{\bibinfo{person}{Wayne~Xin Zhao}, \bibinfo{person}{Kun Zhou}, \bibinfo{person}{Junyi Li}, \bibinfo{person}{Tianyi Tang}, \bibinfo{person}{Xiaolei Wang}, \bibinfo{person}{Yupeng Hou}, \bibinfo{person}{Yingqian Min}, \bibinfo{person}{Beichen Zhang}, \bibinfo{person}{Junjie Zhang}, \bibinfo{person}{Zican Dong}, \bibinfo{person}{Yifan Du}, \bibinfo{person}{Chen Yang}, \bibinfo{person}{Yushuo Chen}, \bibinfo{person}{Zhipeng Chen}, \bibinfo{person}{Jinhao Jiang}, \bibinfo{person}{Ruiyang Ren}, \bibinfo{person}{Yifan Li}, \bibinfo{person}{Xinyu Tang}, \bibinfo{person}{Zikang Liu}, \bibinfo{person}{Peiyu Liu}, \bibinfo{person}{Jian-Yun Nie}, {and} \bibinfo{person}{Ji-Rong Wen}.} \bibinfo{year}{2023}\natexlab{}.
\newblock \showarticletitle{A Survey of Large Language Models}.
\newblock \bibinfo{journal}{\emph{arXiv preprint arXiv:2303.18223}} (\bibinfo{year}{2023}).
\newblock


\bibitem[Zheng et~al\mbox{.}(2023)]%
        {zheng2023judging}
\bibfield{author}{\bibinfo{person}{Lianmin Zheng}, \bibinfo{person}{Wei-Lin Chiang}, {et~al\mbox{.}}} \bibinfo{year}{2023}\natexlab{}.
\newblock \showarticletitle{Judging LLM-as-a-judge with MT-Bench and Chatbot Arena}.
\newblock \bibinfo{journal}{\emph{arXiv preprint arXiv:2306.05685}} (\bibinfo{year}{2023}).
\newblock


\bibitem[Zhu et~al\mbox{.}(2021)]%
        {zhu2021retrievingreading}
\bibfield{author}{\bibinfo{person}{Fengbin Zhu}, \bibinfo{person}{Wenqiang Lei}, \bibinfo{person}{Chao Wang}, \bibinfo{person}{Jianming Zheng}, \bibinfo{person}{Soujanya Poria}, {and} \bibinfo{person}{Tat-S Chua}.} \bibinfo{year}{2021}\natexlab{}.
\newblock \showarticletitle{Retrieving and Reading: A Comprehensive Survey on Open-Domain Question Answering}.
\newblock \bibinfo{journal}{\emph{arXiv preprint arXiv:2101.00774}} (\bibinfo{year}{2021}).
\newblock


\end{thebibliography}

\end{document}